\documentclass{article}

\usepackage{microtype}
\usepackage{graphicx}
\usepackage{booktabs} %

\usepackage{hyperref}

\usepackage[accepted]{icml2025}

\usepackage{amsmath}
\usepackage{amssymb}
\usepackage{mathtools}
\usepackage{amsthm}
\newtheorem{definition}{Definition}[section]

\newtheorem{example}[definition]{Example}

\usepackage[capitalize,noabbrev]{cleveref}
\usepackage{algorithm}
\usepackage{multirow}
\usepackage{xspace}
\usepackage{listings}
\usepackage{graphicx}   %
\usepackage{subcaption} %
\usepackage{enumitem}
\usepackage{array}
\usepackage{wrapfig}
\usepackage{hyperref}
\usepackage{url}
\usepackage{float}

\usepackage{listings}
\usepackage{xcolor}
\newcommand{\tool}{\textsc{Dolphin}\xspace}
\newcommand{\code}[1]{\texttt{\footnotesize #1}}

\newcommand{\cameraready}[1]{\textcolor{orange}{#1}}

\newcommand{\changed}[1]{{#1}}

\lstdefinestyle{PythonStyle}{language=Python,
    basicstyle=\ttfamily\scriptsize,
    breaklines=true,
}

\newcommand{\neusymp}{$P$\xspace}

\newcolumntype{H}{@{}>{\lrbox0}l<{\endlrbox}@{}}  %

\newcommand{\scnote}[2]{{#1}e{#2}}

\theoremstyle{plain}

\theoremstyle{definition}

\theoremstyle{remark}

\icmltitlerunning{\tool{}: A Programmable Framework for Scalable Neurosymbolic Learning}

\begin{document}

\twocolumn[
\icmltitle{\tool{}: A Programmable Framework for Scalable Neurosymbolic Learning}

\icmlsetsymbol{equal}{*}

\begin{icmlauthorlist}
\icmlauthor{Aaditya Naik}{penn}
\icmlauthor{Jason Liu}{penn}
\icmlauthor{Claire Wang}{penn}
\icmlauthor{Amish Sethi}{penn}
\icmlauthor{Saikat Dutta}{cornell}
\icmlauthor{Mayur Naik}{penn}
\icmlauthor{Eric Wong}{penn}
\end{icmlauthorlist}

\icmlaffiliation{penn}{Department of Computer and Information Science, University of Pennsylvania}
\icmlaffiliation{cornell}{Department of Computer Science, Cornell University}

\icmlcorrespondingauthor{Aaditya Naik}{asnaik@seas.upenn.edu}

\icmlkeywords{Machine Learning, ICML}

\vskip 0.3in
]

\printAffiliationsAndNotice{}  %

\definecolor{WowColor}{rgb}{.75,0,.75}
\definecolor{SubtleColor}{rgb}{0,0,.50}
\newcommand{\cmark}{\ding{51}\xspace}%
\newcommand{\xmark}{\ding{55}\xspace}%

\ifdefined\Comment
        \renewcommand{\Comment}[1]{}
\else
        \newcommand{\Comment}[1]{}
\fi
\newcommand{\NA}[1]{\textcolor{SubtleColor}{ {\tiny \bf ($\star$)} #1}}
\newcommand{\Fix}[1]{\textcolor{red}{[#1]}}
\newcommand{\MN}[1]{\textcolor{blue}{#1}}
\newcommand{\LATER}[1]{\textcolor{SubtleColor}{#1}}
\newcommand{\TBD}[1]{\textcolor{SubtleColor}{ {\tiny \bf (!)} #1}}
\newcommand{\PROBLEM}[1]{\textcolor{WowColor}{ {\bf (!!)} {\bf #1}}}

\newcounter{margincounter}
\newcommand{\displaycounter}{{\arabic{margincounter}}}
\newcommand{\incdisplaycounter}{{\stepcounter{margincounter}\arabic{margincounter}}}

\newcommand{\fTBD}[1]{\textcolor{SubtleColor}{$\,^{(\incdisplaycounter)}$}\marginpar{\tiny\textcolor{SubtleColor}{ {\tiny $(\displaycounter)$} #1}}}

\newcommand{\fPROBLEM}[1]{\textcolor{WowColor}{$\,^{((\incdisplaycounter))}$}\marginpar{\tiny\textcolor{WowColor}{ {\bf $\mathbf{((\displaycounter))}$} {\bf #1}}}}

\newcommand{\fLATER}[1]{\textcolor{SubtleColor}{$\,^{(\incdisplaycounter\dagger)}$}\marginpar{\tiny\textcolor{SubtleColor}{ {\tiny $(\displaycounter\dagger)$} #1}}}

\newcommand{\mypara}[1]{\noindent \textbf{#1.}}

\lstset{
    language=Python,
    basicstyle=\ttfamily\small, %
    numbers=left, %
    numberstyle=\tiny\color{gray}, %
    stepnumber=1, %
    numbersep=8pt, %
    backgroundcolor=\color{white}, %
    showspaces=false,
    showstringspaces=false,
    showtabs=false,
    frame=single, %
    rulecolor=\color{black},
    tabsize=4, %
    captionpos=b, %
    breaklines=true, %
    breakatwhitespace=false,
    keywordstyle=\color{blue}, %
    commentstyle=\color{green!60!black}, %
    stringstyle=\color{orange}, %
    morekeywords={as} %
}

\definecolor{ghgreen}{rgb}{0.90,1,0.93}
\definecolor{ghred}{rgb}{1,0.88,0.94}

\definecolor{codegreen}{rgb}{0,0.6,0}
\definecolor{codegray}{rgb}{0.5,0.5,0.5}
\definecolor{codepurple}{rgb}{0.58,0,0.82}
\definecolor{backcolour}{rgb}{0.95,0.95,0.92}

\lstset{  
    language=Java,  
    commentstyle=\color{codegreen},
    keywordstyle=\color{codepurple},
    numberstyle=\tiny\color{codegray},
    stringstyle=\color{blue},
    basicstyle=\footnotesize\ttfamily,
    breakatwhitespace=false,
    breaklines=true,
    captionpos=b,
    keepspaces=true,
    numbers=left,
    numbersep=5pt,
    tabsize=4,
    columns=fullflexible
}

\definecolor{codegreen}{rgb}{0,0.6,0}
\definecolor{codegray}{rgb}{0.5,0.5,0.5}
\definecolor{codepurple}{rgb}{0.58,0,0.82}
\definecolor{backcolour}{rgb}{0.95,0.95,0.92}

\lstdefinestyle{codeqlstyle}{
    commentstyle=\color{codegreen},
    keywordstyle=\color{magenta},
    numberstyle=\tiny\color{codegray},
    stringstyle=\color{codepurple},
    basicstyle=\footnotesize\ttfamily,
    breakatwhitespace=false,         
    breaklines=true,                 
    captionpos=b,                    
    keepspaces=true,                 
    numbers=left,                    
    numbersep=5pt,                  
    showspaces=false,                
    showstringspaces=false,
    showtabs=false,                  
    tabsize=2,
    language=SQL,
    morecomment=[l]{//},%
    morekeywords={select, from, where, and, or, not, predicate, class, extends, import, module, with, without, string,bindingset,if}, %
}

\lstdefinelanguage{MyPrompt}{
  keywords={System, Response, User},
  sensitive = true,
  comment=[l]{//}, 
  morecomment=[s]{/*}{*/}
}
\lstdefinestyle{mypromptstyle}{
    language=MyPrompt,  
    backgroundcolor=\color{backcolour},
    basicstyle=\footnotesize\ttfamily,
    commentstyle=\color{codegreen},
    keywordstyle=\color{codepurple}\bfseries,
    numberstyle=\tiny\color{codegray},
    stringstyle=\color{blue},    
    breakatwhitespace=false,
    breaklines=true,
    captionpos=b,
    keepspaces=true,
    numbers=left,
    numbersep=5pt,
    tabsize=4,
    columns=fullflexible,
    morecomment=[s]{[}{]},
}

\newcommand{\circledsup}[1]{%
  \tikz[baseline=(char.base)]{%
    \node[shape=circle,draw,inner sep=0.5pt] (char) {#1};}%
}

\begin{abstract}
    Neurosymbolic learning enables the integration of symbolic reasoning with deep learning but faces significant challenges in scaling to complex symbolic programs, large datasets, or both. 
    We introduce \tool{}, a framework that tackles these challenges by supporting neurosymbolic programs in Python, executing complex symbolic reasoning on the CPU while vectorizing probabilistic computations and gradient propagation on the GPU.
    Across 13 benchmarks spanning tasks over text, image, and video data, with symbolic reasoning features like recursion and black-box functions, \tool{} converges to state-of-the-art accuracies on the more complex benchmarks while existing frameworks such as Scallop, ISED, and IndeCateR+ fail to converge within the time limit. On simpler benchmarks, \tool{} matches their performance, while achieving these results 1.71x to 62x faster than the baselines. Overall, \tool{} advances the scalability of neurosymbolic frameworks, achieving state-of-the-art efficiency and convergence on difficult benchmarks where existing frameworks struggle. The code is published at \url{https://github.com/Dolphin-NeSy/Dolphin}.
\end{abstract}

\vspace{-0.15in}
\section{Introduction}

\begin{figure}[t!]
    \centering
    \includegraphics[width=0.95\linewidth]{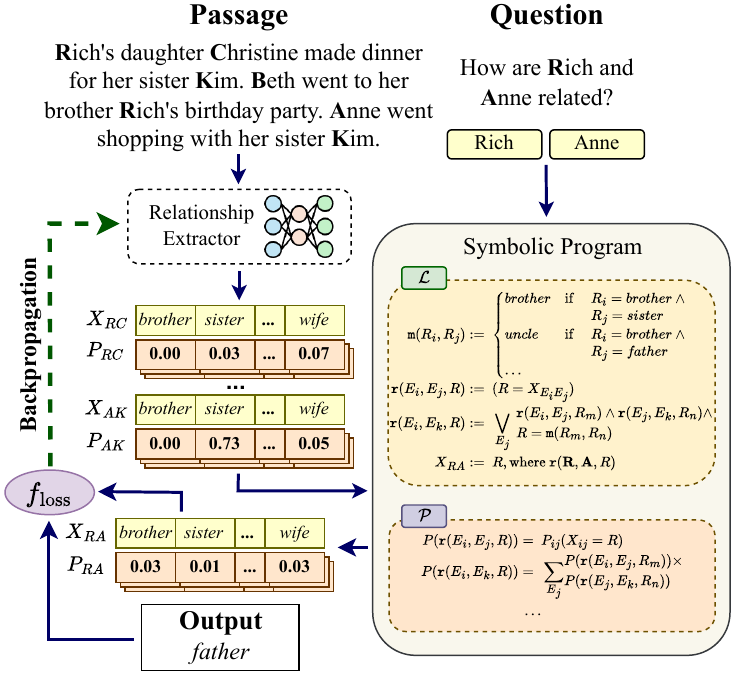}
    \caption{Illustration of the execution of a neurosymbolic program for the kinship reasoning task CLUTRR. While existing neurosymbolic frameworks run the neural models on the GPU, they run the symbolic program entirely on either the CPU (e.g. Scallop) or the GPU (e.g. Logic Tensor Networks), rendering them inefficient in terms of compute and memory, respectively. In \tool{}, both neural models and symbolic programs are specified as PyTorch modules, but only probabilistic computations ($\mathcal{P}$) are vectorized on GPU whereas symbolic computations ($\mathcal{L}$) execute on CPU.}
    \vspace{-0.2in}

    \label{fig:sysarch}
\end{figure}

Deep learning has made great strides in tasks such as image classification, speech
recognition, and natural language processing. With the emergence of foundation models like \mbox{GPT-4} and SAM, deep learning is increasingly applied to more complex tasks. Despite significant strides, these models remain limited in their ability to reliably perform reasoning required for tasks involving structure, logic, and planning, where symbolic approaches traditionally excel~\citep{kambhampati2024llms}.
Neurosymbolic programming~\citep{chaudhuri2021neurosymbolic} has emerged as a promising paradigm to incorporate symbolic reasoning into deep learning models, providing the best of both worlds.

Various frameworks have been developed to improve the programmability and accessibility of neurosymbolic applications~\citep{manhaeve2018deepproblog,li2023scallop,solko2024data}.
These frameworks support complex symbolic reasoning features like recursion and black-box functions, implement efficient differentiable reasoning algorithms, and provide bindings for deep learning frameworks like PyTorch.
However, they incur significant overhead during training.

Figure~\ref{fig:sysarch} shows an example of a kinship reasoning task called CLUTRR whose goal is to infer the relationship between two people based on a passage describing interactions and relationships within a family. A natural neurosymbolic formulation splits this task's computation into a neural component, which extracts relationships from the input passage, and a symbolic component which infers new relationships to obtain the final answer.
The latter involves specifying complex manipulations over symbols ($\mathcal{L}$), e.g., multi-hop kinship reasoning between pairs of family members, and performing probabilistic computations ($\mathcal{P}$) to track the probabilities of the symbols derived using $\mathcal{L}$.
In general, as the complexity of the symbolic program increases, the number of possible results and their associated weights also grows exponentially, leading to a combinatorial explosion in the number of required computations.
This issue is exacerbated by larger datasets usually found in deep learning tasks. Deep learning frameworks typically address this challenge by batching computations across multiple data samples.

Neurosymbolic frameworks like LYRICS~\citep{lyrics} and Logic Tensor Networks (LTN)~\citep{ltn} also batch the computations of both $\mathcal{L}$ and $\mathcal{P}$ on the GPU.
LTN grounds all logits and discrete symbols as tensors, and the aforementioned computations ($\mathcal{L}$ and $\mathcal{P}$) are specified in differentiable first-order logic as operations over those tensors. These programs output a value quantifying the satisfiability of model outputs with respect to logic constraints.
This approach is highly performant for smaller tasks, such as MNIST SumN, where the goal is to predict the sum of $N$ MNIST images.
As we see in Table~\ref{tab:simple_problems} in $\S$~\ref{sec:exp:scalability}, LTN takes around 90 seconds to converge for $N=5$.
However, when the complexity increases to $N=10$, LTN runs out of memory on consumer-grade GPUs (here, with a capacity of 11GBs), due to the combinatorial explosion of required symbols (from $10^5$ to $10^{10}$) and their probabilities that require to be grounded on the GPU.

On the other hand, neurosymbolic frameworks like DeepProbLog~\citep{manhaeve2018deepproblog} and Scallop~\citep{li2023scallop} run neural models on the GPU but use a separate CPU-based backend 
for executing both $\mathcal{L}$ and $\mathcal{P}$.
This avoids issues of memory consumption on the GPU, but the lack of batched computations on CPU results in slowdowns as the problem complexity increases. We see this in MNIST SumN (Table~\ref{tab:all_problems}). Scallop requires around 15 minutes to converge for $N=5$, but needs around 1 and 2 hours to converge for $N=10$ and $N=15$, respectively.

In this paper, we propose \tool as a solution for scaling neurosymbolic learning.  In \tool, we build three key components that effectively tackle scalability challenges with existing neurosymbolic frameworks.
First, we develop a unified representation that efficiently captures the relationships between neural network outputs as PyTorch tensors on GPU and associated discrete symbols as Python objects on CPU. 
Second, we introduce a set of primitives to enable writing symbolic manipulations that can be mapped to computations over these representations, while allowing support for black-box Python functions that simplify the writing of complex symbolic programs.
Third, we develop a set of vectorized {\em 
provenance semirings} \citep{green2007provenance} that are easily pluggable into \tool and enable to efficiently compute symbolic gradients.

Together, these components enable \tool to construct a computation graph that integrates both neural and probabilistic computations ($\mathcal{P}$), ensuring high parallelism and end-to-end differentiability on GPU. At the same time, it runs $\mathcal{L}$ over discrete symbols on CPU, allowing flexible manipulation over arbitrary Python objects. This allows \tool{} to scale effectively to complex problems such as CLUTRR-N (Table~\ref{tab:all_problems}) where $N$ denotes the max length of the reasoning chain in the training dataset.
In the case of Scallop, as the length of the reasoning chain increases, the gap between convergence times dramatically widens.
In contrast, for $N=3$, \tool{} takes around 13 minutes to converge, about 5x faster than Scallop, while for $N=4$, \tool{} takes around 15 minutes, about 8.5x faster than Scallop.
Finally, \tool{} is implemented as a library integrated with PyTorch, allowing users to easily incorporate it into their
existing deep learning pipelines.%

We evaluate \tool{} on a diverse set of neurosymbolic tasks involving text, image, and video, using rich reasoning features like recursion and black-box Python functions.
On simpler problems, neurosymbolic programs written using \tool{} match the accuracy of state-of-the-art methods, while achieving these results 47x, 62x, 8x, and 1.7x faster than baselines like Scallop,  sampling-based frameworks like ISED and IndeCateR+, and solely GPU based methods like LTN respectively.
We also observe that \tool efficiently scales to
more complex benchmarks and larger datasets, achieving state-of-the-art accuracies. While baselines fail to converge on 5 out of 8 such benchmarks within 10 hours, \tool{} requires 5.5 hours in the worst case.

We make the following contributions in this work:
\begin{itemize}[leftmargin=*,itemsep=0.1pt,partopsep=0pt,topsep=0pt]
\item We propose \tool, a novel neurosymbolic programming framework for end-to-end differentiable symbolic reasoning in a scalable manner (\S \ref{sec:technique}).
\item We develop novel Pythonic abstractions and primitives to enable writing complex symbolic manipulations for neurosymbolic programs (\S \ref{sec:syntax}).
\item We design \tool{} to be extendable to new provenances and develop vectorized provenances that can be plugged into \tool for efficient computation of symbolic gradients on parallelizable hardware such as GPU (\S \ref{sec:provenance}).
\item We evaluate \tool{} on a diverse range of 13 challenging neurosymbolic tasks across different domains and show that it effectively scales with increasing problem complexity and dataset size (\S \ref{sec:exp}). 
\end{itemize}
\vspace{-0.1in}

\begin{figure}
\centering
\begin{minipage}{0.9\linewidth}

\begin{lstlisting}[style=PythonStyle, frame=none, escapechar=@]
class SumNNet(torch.nn.Module):
  def __init__(self):
    super(SumNNet, self).__init__()
    self.CNN = MNISTNet()

  def forward(self, imgs):
    d = range(10)
    D_res = Distribution(self.CNN(imgs[0]), d)
    for i in range(1, len(imgs)):
      D_i = Distribution(self.CNN(imgs[i]), d)
      D_res = apply(D_res, D_i, lambda x,y: x + y)
    return get_logits(D_res)\end{lstlisting}\end{minipage}
    \vspace{-0.1in}
        \caption{\tool{} code for the MNIST SumN task.}
        \label{fig:mnist_sum_n_code}
\end{figure}
\begin{figure}[t]
    \centering
    \hfill
    \begin{subfigure}[b]{\linewidth}
        \centering
        \includegraphics[scale=0.4]{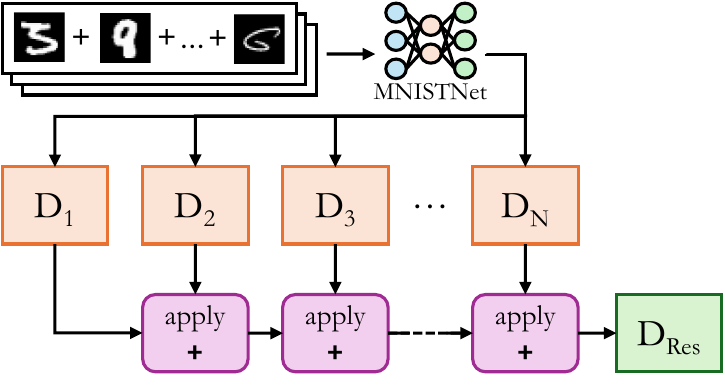}
        \caption{MNIST Sum-N.}
        \label{fig:mnist_graph}        
    \end{subfigure}
    \hfill
    \begin{subfigure}[b]{\linewidth}
    \vspace{0.1in}
        \centering
        \includegraphics[scale=0.4]{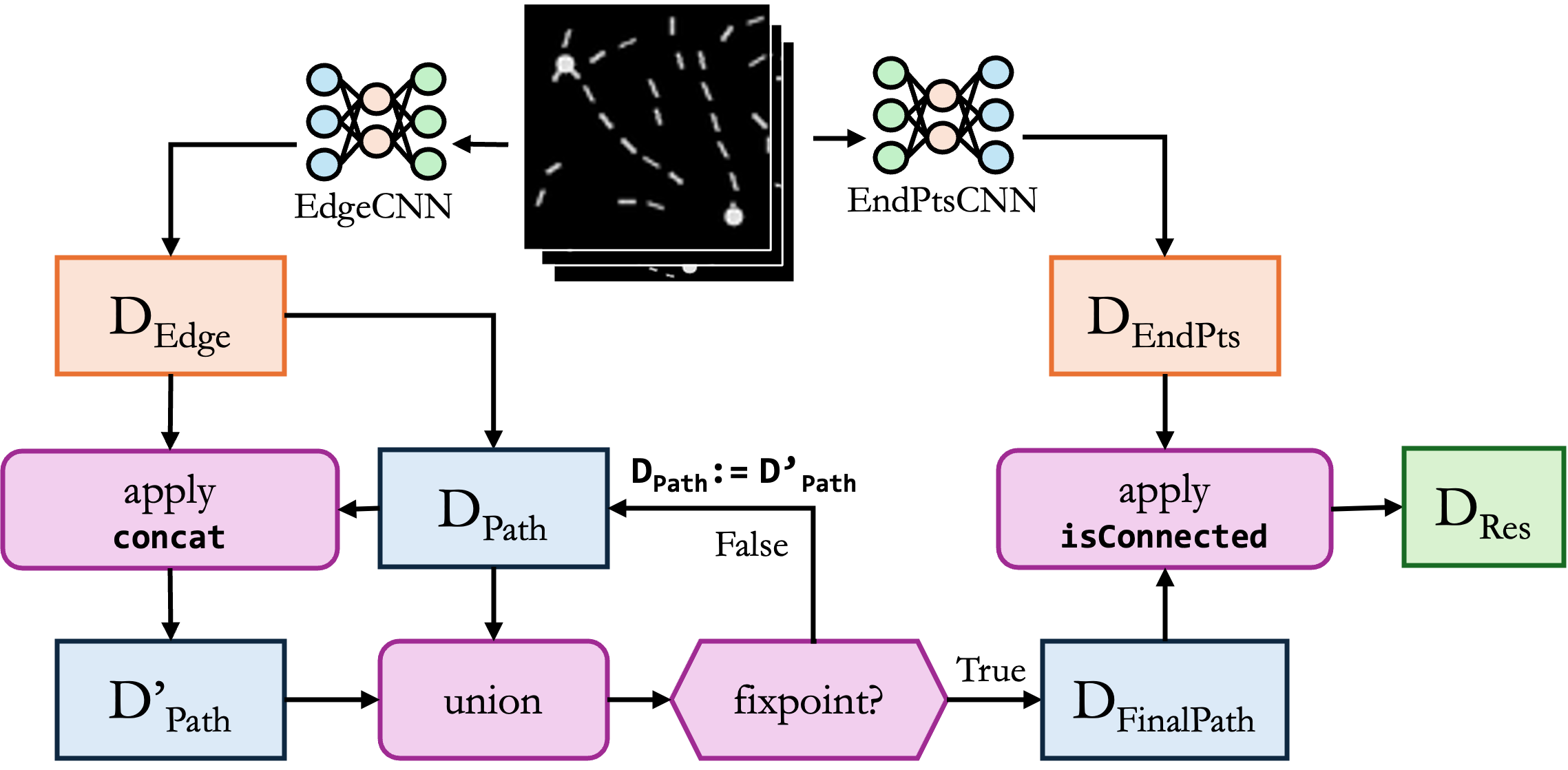}
        \caption{PathFinder.}
        \label{fig:path_graph}
    \end{subfigure}
    \hfill
    \vspace{-0.2in}
    \caption{Computation graphs for two neurosymbolic programs written using \tool{}.}
    \label{fig:computation_graphs}
    \vspace{-0.1in}
\end{figure}

\section{Overview}
\label{sec:overview}

We illustrate \tool{} using the MNIST SumN task \citep{de2024differentiable}, where the goal is to add $N$ MNIST digit images. The task grows exponentially difficult, with $10^N$ possible states and only $9N+1$ labels, making supervision sparse.
Figure~\ref{fig:mnist_sum_n_code} shows the code for this task using \tool{} with PyTorch.
The neural module \code{MNISTNet} is a PyTorch model classifying a batch of images into one of 10 classes representing the digits 0-9.
This is done for each of the $N$ batches of images in the tuple \code{imgs}.
The logits produced by \code{MNISTNet}, representing probability distributions over the digits, are then passed as inputs to the symbolic program.
Lines 8-11 depict a symbolic program written in Python using \tool{} primitives. %

To support training, the symbolic program must track digit probabilities, compute probability distributions over all possibilities ($0$ to $9N$), and propagate gradients for backpropagation. Batched computations further complicate this, making native PyTorch implementations cumbersome.

\tool{} abstracts symbolic computation, letting programmers express logic without handling underlying complexities. 
Lines 8 and 10 of Figure~\ref{fig:mnist_sum_n_code} show how \code{MNISTNet}'s output can be captured within \code{Distribution} objects.
Each \code{Distribution} associates a single collection of digits with the corresponding batch of logits produced by \code{MNISTNet}, along with any gradients and associated metadata.

The programmer can now express the symbolic program in terms of operations over Distributions.
For instance, in line 11, the \code{apply} function is used, taking two Distributions as arguments, along with a lambda function that specifies the addition operation.
Under the hood, \code{apply} combinatorially explores all possible sums of the symbols from \code{D\_res} and \code{D\_i} and calculates their associated probabilities.
The result of \code{apply} is a new \code{Distribution} over the calculated sums, and is stored back into \code{D\_res}.
This is repeated iteratively until all the outputs of the CNN are summed appropriately.

\tool{} provides additional primitives to support more complex symbolic programs.
Figure~\ref{fig:path_graph} shows the computation graph for the PathFinder task~\citep{tay2020long}, which involves recursively building paths to identify if two points in a maze are connected.
The \code{union} primitive is used to support the recursive nature of this program.
Since \code{Distribution} objects associate symbols with the batched logits themselves, probabilistic computations are vectorized and directly operate over PyTorch tensors.

This deep integration of \tool{} into PyTorch allows programmers to write symbolic programs as \textit{symbolic layers} that interact with PyTorch neural layers within a neurosymbolic model.
\tool{} can thus leverage the hardware acceleration supported by PyTorch. %
This contrasts with systems like Scallop, where tensors are converted into Scallop-friendly tags transferred to a process outside the Python environment with CPU-bound probability computations, restricting scalability.

\section{The \tool{} Framework}
\label{sec:technique}

We based \tool's design on four core principles. First is {\em flexible programmability}, to allow writing complex symbolic manipulations ($\mathcal{L}$) with Python's rich and expressive language features.
Second, probabilistic computations ($\mathcal{P}$) must allow {\em end-to-end differentiability} on the GPU. Third, \tool{} must be {\em scalable} to tasks with large data and problem complexity. Finally, it must be {\em tunable}, allowing developers to define and choose provenances, treating them like deep learning hyperparameters.

Together, these principles help address the challenges of scaling neurosymbolic frameworks. Flexible programmability and tunability allow us to write complex neurosymbolic programs, while GPU differentiability and scalability work towards tackling problem and data complexity.
We now describe \tool{} and show how we realize these principles.

\subsection{The \tool Syntax}
\label{sec:syntax}

\looseness=-2
To allow flexible programmability, \tool provides an interface that developers can use to express symbolic programs in a Pythonic manner.

\begin{figure}[t]
\footnotesize
\begin{subfigure}[c]{0.51\linewidth}
\centering
\[
\begin{array}{rcrl}
   \text{Symbol} & :: & s & \in \ \ S \ \ \text{ (\code{objects})} \\
    \text{Tag} & :: & t & \in \ \ T \ \ \text{ (\code{tensors})}  \\
    \text{Distribution} & :: & D & \in \ \ \mathbb{D} = S \rightarrow T %
\end{array}
\]

\end{subfigure}
\begin{subfigure}[c]{0.51\linewidth}
\centering
\[
\begin{array}{rcl}
     \textsc{Apply} &:& \mathbb{D}^K \times (S^K \rightarrow S) \rightarrow \mathbb{D} \\
      \textsc{Filter} &:& \mathbb{D} \times (S \rightarrow \mathbb{B}) \rightarrow \mathbb{D} \\
      \textsc{ApplyIf} &:& \mathbb{D}^K \times (S^K \rightarrow S) \times (S^K \rightarrow \mathbb{B}) \rightarrow \mathbb{D} \\ 
      \textsc{Union} &:& \mathbb{D} \times \mathbb{D} \rightarrow \mathbb{D} \\ 
      \textsc{GetProbs} &:& \mathbb{D} \rightarrow [0,1]^N
\end{array}
\]

\end{subfigure}
\caption{Formal definition of \tool{}'s programming abstractions (top) and primitives (bottom).}
\vspace{-.15in}
\label{fig:syntax}
\end{figure}

\subsubsection{Abstractions}
\tool provides three main abstractions for expressing symbolic programs, shown in Figure~\ref{fig:syntax}.
\emph{Symbols} $S$ represent symbolic entities relevant to the program.
These entities can be any Pythonic object, such as hand-written digits in MNIST-SumN or coordinates of points in PathFinder.
\emph{Tags} $T$ are tensors that represent their likelihoods.
Typically, tags for symbols are derived from the outputs of machine learning models, such as the logits produced by the CNN classifier in MNIST-SumN.
Finally, \emph{Distribution} $D$ maps a collection of symbols to their corresponding tags.

Distributions serve as the fundamental datatype of a \tool program and act as its main interface with a PyTorch model. As seen in the following code snippet from Figure~\ref{fig:mnist_sum_n_code}:

\vspace{-0.2in}
{\footnotesize
\begin{equation*}
    \code{$\texttt{D}_{\text{res}}$ = Distribution(self.CNN(imgs[0]), d)}
\end{equation*}
}

\vspace{-0.1in}
\noindent
the logits output by the model are directly passed to the Distribution object, effectively acting as an input to the symbolic program.
These logits form the batched \textit{tags} within a Distribution object which also maintains the set of corresponding symbols \texttt{d}.
The symbolic manipulations in a \tool{} program occur over the discrete symbols, while the probabilistic computations occur over tags stored as PyTorch tensors.
This enables a seamless integration between the PyTorch model and the symbolic program.

This has several advantages.
First, it preserves the gradients of the model output throughout the symbolic program, enabling end-to-end differentiability via PyTorch's autograd, addressing the second core principle of \tool.
Second, it allows \tool to perform operations over an entire batch of tags as per the principle of scalability, leveraging the vectorized operations provided by PyTorch.
\tool can thus operate efficiently on specialized hardware like GPUs, allowing the symbolic program to scale effectively.
Third, since symbol and tag computations are effectively decoupled, operations over symbols can be run on the CPU, allowing the support for arbitrary Python objects and functions in the symbolic program, even while the tag computations are performed on GPU.
\begin{example}
\label{eg:distribution}
    Consider MNIST images $I_1, I_2$ for the SumN task discussed in Section~\ref{sec:overview}. Let $f_\text{CNN}$ be the neural model which classifies image $I_j$ into one of 10 classes representing digits 0 to 9.
    Let $f_\text{CNN}(I_1) = \{0.00, 0.90, \ldots, 0.01\}$ and $f_\text{CNN}(I_2) = \{0.78, 0.09, \ldots, 0.00\}$.
    We thus define the following Distributions:
    \begin{enumerate}
        \vspace{-0.1in}
        \item $D_1 = \{0 \rightarrow 0.00, 1 \rightarrow 0.90, \ldots, 9 \rightarrow 0.01\}$
        \item $D_2 = \{0 \rightarrow 0.78, 1 \rightarrow 0.09, \ldots, 9 \rightarrow 0.01\}$
    \end{enumerate}
\end{example}

\subsubsection{Operations}
\label{sec:operations}
\tool{} provides five operations to allow the expression of complex neurosymbolic programs in conjunction with user-defined functions, shown in 
Figure~\ref{fig:syntax}.
\vspace{-0.1in}
\paragraph{\textsc{Apply}.}
This is the primary operation that can be used to manipulate Distributions.
It takes as inputs $K \geq 1$ Distributions, along with a function $f$ of the same arity.
This function defines operations over the symbols of $K$ distributions.
\textsc{Apply} then computes the results of $f$ over all possible combinations of arguments sourced from the symbols of
the Distributions as well as their associated tags, and returns a new Distribution with these results and tags.
This occurs in two stages akin to the popular map-reduce pattern.
In the \textit{map} stage, \textsc{Apply} computes the results of $f$ over the symbols of the input Distributions and conjuncts their tags:

\vspace{-0.2in}
{\footnotesize
\begin{equation}
\begin{split}
R\ =\ & \{\ (f(s_1, s_2, \ldots, s_k), (t_1 \otimes t_2 \otimes \ldots \otimes t_k))\ | \\
      & D_i(s_i) = t_i, i = 1, \ldots, k\ \}
\end{split}
\label{eq:eval}
\end{equation}
}

\vspace{-0.1in}
Here, the tag of each result symbol $f(s_1, s_2, \ldots, s_k)$ is the conjunction $\otimes$ of the tags $(t_1, t_2, \ldots, t_k)$ of the input symbols it was derived from.
The function $f$ is executed sequentially on the CPU for each combination of symbols as function $f$ can be any user-defined Python function, including complex control flows and operations like regex parsing, image processing, or Python's \code{eval()}.
It may also be a many-to-one function and the tags shared by a resulting symbol must be aggregated
to form the final tags of the output Distribution.
We, therefore, \textit{shuffle} the results from the map stage to compute a function $M$ from each symbol to tags from $R$ associated with it:

\vspace{-0.15in}
{\footnotesize
\begin{equation}
M = \lambda \ s\ .\ \{\ t\ |\ (s, t) \in R \ \}
\end{equation}
}

\vspace{-0.1in}
We then proceed to the \textit{reduce} stage, where we aggregate the tags of each symbol in $M$ using disjunction $\oplus$ to produce the final Distribution $D_\text{res}$:

\vspace{-0.1in}
{\footnotesize
\begin{equation}
D_\text{res} = \lambda\ s\ .\ \bigoplus\ \{\ t\ |\ t \in M(s) \ \}
\label{eq:agg}
\end{equation}
}

\vspace{-0.1in}
Since the tags here are PyTorch tensors representing probabilities, the implementations of the conjunction and disjunction
operations are dictated by the underlying provenance specified by the program, detailed in Section~\ref{sec:provenance}.

\begin{example}
\label{eg:apply}
    Continuing from Example~\ref{eg:distribution}, let function $f(x, y) = x + y$ be applied to $D_1$ and $D_2$ to produce a new Distribution $D' = \texttt{apply}(D_1, D_2, f)$.
    $D'$ thus represents a Distribution over the sum of the symbols from $D_1$ and $D_2$:
    $$D' = \{ 0 \rightarrow 0.00, 1 \rightarrow 0.70, \ldots, 18 \rightarrow 0.00\}$$
    Consider the tag of $D'(1)$. 1 can be a result of $D_1$ being 0 and $D_2$ being 1, or of $D_1$ being 1 and $D_2$ being 0:
    \begin{align*}
        D'(1) & = (D_1(0) \otimes D_2(1)) \oplus (D_1(1) \otimes D_2(0)) \\
        & = (0.00 \otimes 0.09) \oplus (0.90 \otimes 0.78)
    \end{align*}
    This expression's value (e.g. 0.70) depends on the provenance specified (e.g. DAMP), discussed in Section~\ref{sec:provenance}.
\end{example}

\mypara{\textsc{Filter}}
The \textsc{Filter} operation is used to filter out symbols from a Distribution.
It takes in a single Distribution, along with a user-defined function that returns a boolean value.
This operation then returns a new Distribution that contains only symbols that satisfy the condition with their tags.
\begin{example}
\label{eg:filter}
    Continuing from Example~\ref{eg:distribution},
    assume we want a Distribution over just the even symbols of $D_1$.
    We can consider a filtering function $f(x) = (x \mod 2 == 0)$. The resulting Distribution will have all the odd-numbered symbols completely removed:
    \begin{align*}
        D' & = \texttt{filter}(D_1, f) \\
           & = \{0 \rightarrow 0.00, 2 \rightarrow 0.02, \ldots, 8 \rightarrow 0.01\}
    \end{align*}
\end{example}

\mypara{\textsc{ApplyIf}}
This operation is a conditional version of \textsc{Apply}.
It takes in $K$ Distributions and functions $f_{\textit{apply}}$ and $f_{\textit{cond}}$ of the same arity.
For each combination of symbols from the $K$ Distributions, \textsc{ApplyIf} computes $f_{\textit{apply}}$ and
its associated tags only if the condition $f_{\textit{cond}}$ is satisfied over that combination of symbols.
The operation then returns a new Distribution with these results and tags.

\mypara{\textsc{Union}}
The \textsc{Union} operation takes in two Distributions and returns a new Distribution containing the union of the input symbols, along with their tags.
Any symbols common to both input Distributions have their tags disjuncted.
\begin{example}
    Consider Distributions $D_1 = \{0 \rightarrow 0.01, 1 \rightarrow 0.24\}$ and $D_2 = \{0 \rightarrow 0.63, 4 \rightarrow 0.37\}$.
    The union will be:
    $$\texttt{union}(D_1, D_2) = \{0 \rightarrow 0.64, 1 \rightarrow 0.24, 4 \rightarrow 0.37\}$$
\end{example}

\mypara{\textsc{GetProbs}}
The \textsc{GetProbs} operation extracts the probabilities from the tags of a Distribution.
This is used mainly once the symbolic program has been executed to extract the final probabilities of the symbols in the output Distribution.
These probabilities can then be used to compute the loss function for training the neural model.

\subsubsection{Writing Complex \tool{} Programs}

Some neurosymbolic tasks require the writing of programs containing complex control flows and recursion (e.g. the PathFinder task, $\S$~\ref{sec:exp:benchmarks}). This can be done in one of two ways. The simplest way is to specify any control flow operations within the user-defined functions supplied to the \tool{} operations. Alternatively, one can specify branches of control flow separately and merge their results  via \textsc{Union}, as shown in Figure~\ref{fig:transitive_closure_app} (Appendix~\ref{app:control_flow} for more details).

In some cases, even though \tool{} limits the effect of combinatorial explosion in terms of efficiency and memory usage, the number of combinations may still be excessive. In such cases, \tool{} allows developers to sample subsets of symbols from Distribution objects specified within the symbolic program, effectively limiting the number of symbols processed in each operation.

\subsection{\tool Provenances}
\label{sec:provenance}

The \tool primitives discussed above define how to conjunct or disjunct tags corresponding to the symbol manipulations, e.g. Equations (\ref{eq:eval}) and (\ref{eq:agg}).
These tag operations are achieved by using a mathematical framework called \textit{provenance semirings}~\citep{green2007provenance}.
Provenance semirings provide generalized algebraic structure to propagate probabilities over tagged data.

Designing and implementing provenances can be challenging since they must be accurate enough to capture the semantics
of the symbolic program, while at the same time being coarse enough to maintain computational feasibility.
Furthermore, the provenances must be differentiable.

While neurosymbolic frameworks like Scallop~\citep{li2023scallop} implement differentiable provenances, they
are not designed to leverage hardware accelerations or batched optimizations due to the CPU-bound nature of their implementations.
Frameworks like LTN use t-norms that are more amenable to vectorization, but lack support for more complex provenances such as Differentiable Top-$k$ Proofs (DTKP)~\cite{huang2021scallop}.
We thus design differentiable, vectorized provenances in \tool{} to enable GPU support.

We simplify the definition of provenances as a 5-tuple: $(T, \mathbf 0, \mathbf 1, \otimes, \oplus)$.
Here, $T$ is the tag space, $\otimes: T \times T \to T$ is the conjunction operator with identity $\mathbf 0$,
and $\oplus: T \times T \to T$ is the disjunction operator with identity $\mathbf 1$.
We then implement two differentiable provenances in \tool{}: Differentiable Add-Mult Probabilities (DAMP) and
Differentiable Top-K Proofs (DTKP).
Table~\ref{tab:prov-torch} summarizes the operations of these provenances.
While building the neurosymbolic program, the developer may specify which provenance to use, satisfying the core principle of tunability.

\begin{table*}
    \centering
    \caption{\tool{} provenances implemented in PyTorch.}
    \setlength{\tabcolsep}{0.2em}
    \scriptsize
    \resizebox{\textwidth}{!}{
    \begin{tabular}{l | c| c| c| c |c}
        \toprule
        \textbf{Provenance} & \textbf{Domain} & $\mathbf{0}$ & $\mathbf{1}$ & $t\oplus t'$ & $t\otimes t'$ \\
        \midrule
        DAMP & $[0,1]$ & 0 & 1 & $\text{clamp}_0^1(t+t')$ & $t \cdot t'$ \\
        DTKP-AM & $[0,1]\cup\{\infty, -\infty\}$  & $\mathbf{\hat 0}_{ij} = -\infty$ &
         $\mathbf{\hat 1}_{ij} = \begin{cases}
            \infty & i=1 \\
            -\infty & i>1
        \end{cases}$ & $\text{top}_k(\text{cat}(t, t'))$ & $\text{top}_k([\min(|t_i|,|t'_j|) \mid (t_i,t'_j)\in t\times t'])$ \\
        \bottomrule
    \end{tabular}
    }
    \label{tab:prov-torch}
    \vspace{-0.17in}
    
\end{table*}

\looseness=-1
\mypara{Differentiable Add-Mult Probabilities}
Differentiable Add-Mult Probabilities (DAMP) is a popular technique that uses the probability space as its tag space: $T=[0,1]$.
Its conjunction operation $\otimes$ is defined as the product of probabilities, clamped at $\mathbf{1}$,
and its disjunction operation $\oplus$ is defined as the sum of probabilities.
The main assumption underlying the DAMP operations is that the input Distributions are mutually exclusive and
independent.
This assumption allows DAMP to compute probabilities extremely efficiently, as the operations are simple and can be easily
vectorized.

\mypara{Differentiable Top-$k$ Proofs}
Differentiable Top-$k$ Proofs (DTKP)~\citep{huang2021scallop} was proposed to overcome the shortcomings of DAMP.
This provenance tracks a set of up to $k$ \textit{proofs} for each symbol.
Each proof denotes the set of input symbols necessary to derive the output symbol.
These proofs are then used to compute the probabilities of the output symbols.
In Scallop, DTKP tags are converted into probabilities via differentiable weighted model counting (WMC).
This form of DTKP, which we call DTKP-WMC, is computationally hard and is by nature difficult
to vectorize due to the varying sizes of proof sets and the WMC procedure.
We hence design a novel vectorized approximation of DTKP-WMC; we term DTKP-AM (DTKP with Add-Mult), that can be efficiently computed on the GPU.

We first define the structure of tags in DTKP-AM to conform to the constraints of PyTorch tensors.
Each tag $t$ for a symbol $s$ is a 2-dimensional tensor of shape $(k, |I|)$, where $k$ is the maximum number of proofs to be
retained and $I$ is an ordered list of all \textit{input symbols} (symbols that are present in the input Distributions).
Each row $t_i$ of $t$ corresponds to one of the tag's $k$ proofs.
Each element $t_{ij}$ thus represents the probability of the $j$th input symbol in the $i$th proof:

\vspace{-.2in}
{\footnotesize
\[
t_{ij} = \begin{cases}
    p_{j} & \text{if the $j$th symbol is present in the $i$th proof} \\
    \mathbf{\hat 0}_{ij} & \text{otherwise}
\end{cases}
\]
}

\vspace{-0.15in}
where $p_{j}$ is the probability of the $j$th input symbol.
The probability of each proof is then computed by taking the product of the normal:

\vspace{-.25in}
{
\footnotesize
\[
    \Pr(t_i) = \prod_j \text{norm}(t_{ij}), \  \text{where} \ 
    \text{norm}(t_{ij}) = \begin{cases}
        1 & t_{ij} = +\infty \\
        0 & t_{ij} = -\infty \\
        t_{ij} & \text{otherwise}
    \end{cases}
\]
}

\vspace{-0.1in}
We next define the operations of DTKP-AM in Table~\ref{tab:prov-torch}.
The $\oplus$ operation is defined as the union of two tag tensors $t$ and $t'$ while $\otimes$ is defined as the element-wise minimum of the normalized elements of all possible combinations of proofs in $t$ and $t'$.
In each case, the $\text{top}_k$ operation retains only up to $k$ proofs with the highest probabilities.

These definitions thus allow us to take advantage of the benefits of the DTKP provenance while enabling efficient computation
on the GPU.
To calculate the probability of the entire tag, DTKP-AM adds the probabilities of the individual proofs and clamps it at 1. We provide a detailed discussion of DTKP-AM in Appendix~\ref{app:dtkp}.

\subsection{Building the \tool Program}
The programmer specifies the neurosymbolic task using a Python program \neusymp, which integrates neural components with symbolic operations via \tool's interface. Given a dataset $\mathcal{D}$ and one or more neural networks $M_1, \ldots, M_k$, \tool constructs a computation graph where symbolic transformations occur on the CPU, and probabilistic computations, including neural network inference, are efficiently executed on the GPU. All computations leverage distribution objects $D_i$, enabling end-to-end differentiability and scalability. Training optimizes the objective function
\vspace{-0.05in}
$$\phi(\theta) = \min_{\theta} \sum_{(x,y)\in \mathcal{D}}  \mathbb{L}(\text{\neusymp}(M_{\theta}(x)), y),$$
where $\mathbb{L}$ is the loss function (e.g., binary cross entropy).

\vspace{-0.1in}
\section{Experiments}
\label{sec:exp}

We evaluate \tool on a set of 13 benchmarks of varying complexity and scale across 5 neurosymbolic tasks. Our evaluation addresses the following research questions:
\begin{itemize}[leftmargin=*,itemsep=1pt]
    \item \textbf{RQ1: Scalability.} Can \tool{} scale to tasks and datasets beyond the scope of existing SOTA frameworks?
    \item \textbf{RQ2: Accuracy.} Do models written in \tool{} converge to SOTA accuracies in less training time?
    \item \textbf{RQ3: Provenance Comparisons.} Which provenances are most effective for each benchmark?
\end{itemize}

\subsection{Benchmarks} 
\label{sec:exp:benchmarks}
We describe the benchmarks used to evaluate \tool and give additional information about the experiment setup and \tool{} code for each benchmark in Appendix~\ref{app:code}.

\mypara{MNIST SumN}
The MNIST SumN (or briefly, SumN) task from \citep{de2024differentiable} takes as inputs $N$ handwritten digits from the MNIST dataset
and returns their sum.
We consider three versions of this task: \textbf{SumN-5} ($N=5$), \textbf{SumN-10} ($N=10$), and \textbf{SumN-15} ($N=15$).

\mypara{Hand-Written Formula (HWF)}
The HWF task from~\cite{li2020closed} takes as input a set of images of handwritten digits and arithmetic operators representing
a formula.
The task is to evaluate the formula and return the result.
We consider three versions based on formula length: \textbf{HWF-7} (up to 7), \textbf{HWF-15} (up to 15),
and \textbf{HWF-19} (up to 19).

\mypara{PathFinder}
PathFinder (or Path)~\citep{tay2020long} tests the ability of an agent to reason over long-range dependencies within an image of two dots and a sequence of curved and dashed lines. The task is to identify whether the two dots are connected via the lines.
We consider three versions based on the image size in pixels: \textbf{Path-32} (32 x 32), \textbf{Path-128}
(128 x 128), and \textbf{Path-256} (256 x 256).

\mypara{CLUTRR}
In this task from~\cite{sinha2019clutrr}, given some text containing information about several individuals and some of their relationships, the model must infer the relationship between two given individuals, which is not explicitly provided in the input. We consider two versions, where the training data contains relation chains of lengths up to 3 (\textbf{CLUTRR-3}) or 4 (\textbf{CLUTRR-4}).

\mypara{Mugen}
In this task from~\cite{hayes2022Mugen}, given a 3.2 second long video of gameplay footage and text captioning the video, the goal is to measure how aligned the text is with the video.
There are two variants: Mugen-TVR, where the model retrieves the video that best aligns with the text, and Mugen-VTR, where the model retrieves the text that best aligns with the video.
We consider two versions of this task: \textbf{1K} and \textbf{5K} comprising 1000 and 5000 training samples.

\subsection{Experimental Setup and Baselines}
\mypara{Setup}
All experiments, except CLUTRR, were run on machines with two 20-core Intel Xeon Gold 6248
CPUs, four NVIDIA GeForce RTX 2080 Ti (11 GB) GPUs, and 768 GB RAM.
Since CLUTRR demands more GPU memory due to running the RoBERTa model with a standard batch size of 16, all programs for this benchmark were run with a NVIDIA A100 40GB GPU.
We ran each tool thrice until convergence or until a soft timeout of 10 hours was reached and report the average best accuracy and training time. For HWF/MNIST, we use the same CNN architecture as Scallop (Appendix \ref{app:code}). For CLUTRR, we use Scallop’s Roberta configuration: a pretrained model (roberta-base) finetuned while training the classification head.

\mypara{Baselines}
We select Scallop~\citep{li2023scallop}, a contemporary state-of-the-art neurosymbolic framework supporting differentiable programming optimized to run on the CPU in parallel using multiple cores.
We also choose two sampling-based gradient approximation methods, ISED~\citep{solko2024data} and IndeCateR+~\citep{de2024differentiable}.
We also include Logic Tensor Networks (LTN)~\citep{serafini2016logic}, which combines first-order logic with continuous optimization by compiling logical constraints into a computation graph on the GPU.
We compare \tool against Scallop on all benchmarks, and against ISED and IndeCateR+ on SumN and HWF. We compare SumN with LTN, but were unable to write HWF in LTN (explained in Appendix~\ref{app:ltn-no-hwf}). Since LTN also runs out of memory for a simpler benchmark like SumN-10, we do not compare against it for the more complex benchmarks of Path, CLUTRR, and Mugen. Similarly, we do not evaluate ISED and IndeCateR+ on these benchmarks, as ISED already fails to scale for simpler tasks like SumN-10 and HWF-7, while IndeCateR+ does not support their recursive structures.

\begin{table}
    \centering
    \caption{Training times (in seconds) for \tool{} and Scallop on all benchmarks. Training times more than 10 hours are highlighted in red. The scaling factor $\alpha$ is the ratio of the total training times of Scallop to \tool.}
    \footnotesize
    \setlength{\tabcolsep}{1em} %
    \begin{tabular}{l c cc}
    \toprule
    \textbf{Task} 
      & \textbf{\tool{}} 
      & \multicolumn{2}{c}{\textbf{Scallop}} \\
     & $T_{\text{total}}$ 
     & $T_{\text{total}}$ & $\alpha$\\
    \midrule
    SumN-5
      & 53.86 
      & 923.78 & 17.15 \\
    SumN-10 
      & 104.91 
      & \scnote{3.42}{3} & 32.56 \\
    SumN-15
      & 157.05 
      & \scnote{7.41}{3} & 47.18 \\
    \midrule
    HWF-7 
      & \scnote{2.45}{3} 
      & \scnote{9.99}{3} & 4.08 \\
    HWF-15 
      & \scnote{9.78}{3} 
      & \textbf{\textcolor{red}{\scnote{1.66}{5}}} & 16.97 \\
    HWF-19 
      & \scnote{1.63}{4} 
      & \textbf{\textcolor{red}{\scnote{1.82}{5}}} & 11.16 \\
    \midrule 
    Path-32 
      & \scnote{1.29}{4} 
      & \scnote{2.2}{4} & 1.71  \\
    Path-128 
      & \scnote{1.67}{4} 
      & \textbf{\textcolor{red}{\scnote{4.17}{4}}} & 2.49 \\
    Path-256 
      & \scnote{1.97}{4} 
      & \textbf{\textcolor{red}{\scnote{1.14}{5}}} & 5.78 \\
    \midrule
    CLUTRR-3
      & 807.12
      & \scnote{4.29}{3} & 5.32  \\
    CLUTRR-4
      & 923.86
      & \scnote{7.83}{3} & 8.48  \\
    \midrule
    Mugen-1K 
      & \scnote{2.39}{3} 
      & \scnote{6.71}{3} & 2.81  \\
    Mugen-5K
      & \scnote{1.15}{4} 
      & \scnote{3.59}{4} & 3.12 \\
    \bottomrule
    \end{tabular}
    \label{tab:all_problems}
    \vspace{-0.2in}
\end{table}

\begin{table*}[t]
    \centering
    \caption{Training time (in seconds) for \tool{}, LTN, ISED, and IndeCateR+ on SumN and HWF. Training times more than 10 hours are highlighted in red. $\alpha$ is the ratio of training times of the baselines to \tool{}. OOM occurred on an NVIDIA GeForce RTX 2080 Ti (11 GB).}
    \footnotesize
    \setlength{\tabcolsep}{1em} %
    \begin{tabular}{l c cc cc cc}
    \toprule
    \textbf{Task} 
      & \textbf{\tool{}} 
      & \multicolumn{2}{c}{\textbf{LTN}} 
      & \multicolumn{2}{c}{\textbf{ISED}} 
      & \multicolumn{2}{c}{\textbf{IndeCateR+}} \\
     & $T_{\text{total}}$ 
     & $T_{\text{total}}$ & $\alpha$
     & $T_{\text{total}}$ & $\alpha$
     & $T_{\text{total}}$ & $\alpha$ \\
    \midrule
    SumN-5
      & 53.86 
      & 92.54 & 1.72 
      & 299.63 & 5.56 
      & 416.78 & 7.74 \\
    SumN-10 
      & 104.91 
      & \textbf{\textcolor{red}{OOM}} & - 
      & \scnote{2.17}{3} & 20.64 
      & 385.65 & 3.68 \\
    SumN-15
      & 157.05 
      & \textbf{\textcolor{red}{OOM}} & - 
      & \scnote{9.8}{3} & 62.41 
      & 548.28 & 3.49 \\
    \midrule
    HWF-7 
      & \scnote{2.45}{3} 
      & \multicolumn{2}{c}{N.A.} 
      & \scnote{4.02}{3} & 1.64
      & \scnote{1.35}{4} & 5.51  \\
    HWF-15 
      & \scnote{9.78}{3} 
      & \multicolumn{2}{c}{} 
      & \scnote{2.31}{4} & 2.36
      & \scnote{2.51}{4}  & 2.57  \\
    HWF-19
      & \scnote{1.63}{4} 
      & \multicolumn{2}{c}{} 
      & \textbf{\textcolor{red}{\scnote{9.34}{4}}} & 5.73 
      & \textbf{\textcolor{red}{\scnote{6.27}{4}}} & 3.85 \\
    \bottomrule
    \end{tabular}
    \vspace{-.05in}
    \label{tab:simple_problems}
\end{table*}
\subsection{RQ1: Scalability}
\label{sec:exp:scalability}

Table \ref{tab:all_problems} presents the total training times ($T_{\text{total}}$) till convergence in seconds for \tool and Scallop across all benchmarks, alongside the scaling factor $\alpha$ (the ratio of the total training times of the baselines to \tool). Table~\ref{tab:simple_problems} shows the same for the remaining baselines over
SumN and HWF. We set a soft timeout of 10 hours, though we still report the training times that run over highlighted in red. The results demonstrate that \tool{} advances the state-of-the-art in neurosymbolic learning by scaling to more complex problems, e.g., larger versions of HWF, Path, and Mugen, that are beyond the reach of the other baselines which time out within 10 hours. Even for the other benchmarks where baselines do not time out, \tool{} achieves a scalability factor of up to 47x, 62x, and 3.49x against Scallop, ISED, and IndeCateR+ with an average speed up of 13.95x across all baselines for all benchmarks.

We also use the training times per epoch to calculate a scalability factor $\alpha_\text{epoch}$ (Table~\ref{tab:scalability}, Appendix~\ref{app:code}). We see that among cases where baselines timeout, $\alpha_\text{epoch}$ is up to 280x for HWF-19 and is 40.6x faster on average. This results in \tool{} effectively training for more epochs in less time compared with the other baselines, which also allows it to converge to higher accuracies, as we see in Figure~\ref{fig:accuracy}.
We expand on these results in the next RQ.

\subsection{RQ2: Accuracy}

\begin{figure*}[b!t]
    \centering
    \includegraphics[scale=0.4]{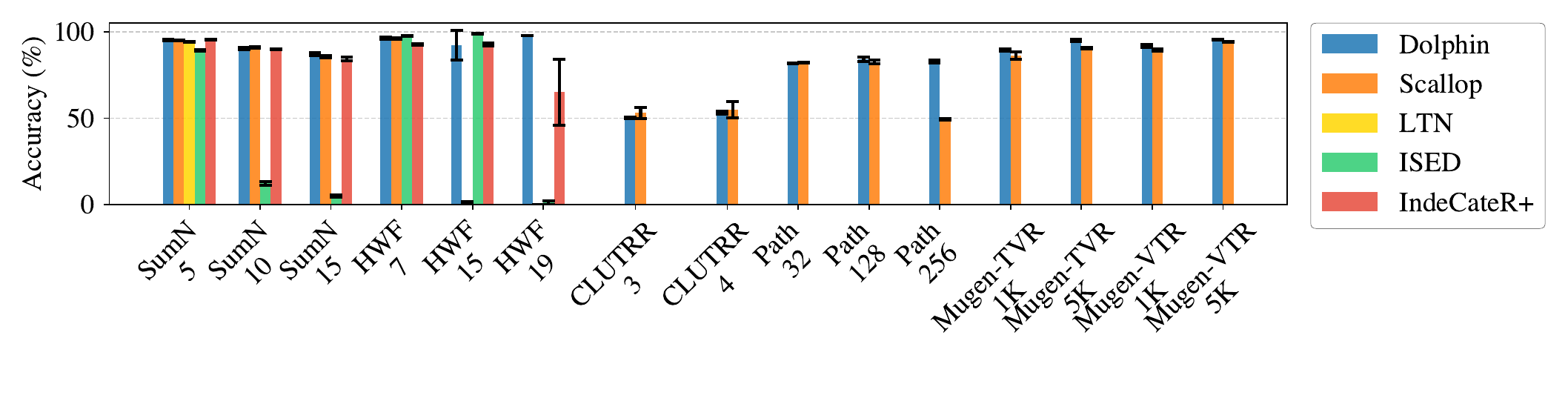}
     \vspace{-0.3in}
    \caption{Accuracy of \tool and baselines trained for up to 10 hours across all benchmarks.}   
    \label{fig:accuracy}
     \vspace{-0.2in}
    
\end{figure*}

Figure \ref{fig:accuracy} presents the accuracy of \tool and the baselines on the different benchmarks trained for up to 10 hours.
\tool accuracies are marked in blue.
In all cases, for \tool, we report the accuracies of the best-performing provenance.
We use the DAMP provenance for MNIST, CLUTRR, and Mugen benchmarks, and the DTKP-AM provenance for the HWF and PathFinder benchmarks.

We observe that in all cases, \tool achieves state-of-the-art accuracy among general-purpose neurosymbolic frameworks, except in CLUTRR, where
\tool's accuracy is slightly lower than Scallop's. As we scale up to larger versions of the benchmarks, \tool achieves better accuracy, because the baselines either report lower accuracy due to the complexity of the benchmark (e.g., black-box sampling techniques such as ISED on HWF) or fail to converge within 10 hours due to slower per-epoch train time (e.g., Scallop on PathFinder-256, IndeCateR+ on HWF-19). As a result, \tool significantly outperforms the second-best benchmark on the largest dataset versions, achieving up to a 20\% gain on HWF and 33\% on Path.
In some cases, given no timeout, Scallop and IndeCateR+ eventually converge to accuracies comparable to \tool{}, as we show in Appendix~\ref{app:acc_no_timeout}. However, doing so requires significantly more training time, as discussed earlier in Section~\ref{sec:exp:scalability}.

These results show that not only do \tool's scalability improvements not come at the cost of accuracy, but \tool enables SOTA accuracy when previously unattainable.

\subsection{RQ3: Provenance Comparisons}

We perform ablation studies to compare the effectiveness of the DAMP and DTKP-AM provenances for each benchmark.
We share the graphs in Figure~\ref{fig:comparison_prov} (Appendix~\ref{app:prov_compare}). In all cases, training with the DAMP provenance takes around 24.19 seconds per epoch less than with DTKP-AM on average.

However, the effectiveness of each provenance varies across benchmarks. For all variations of Path, CLUTRR, and Mugen, both provenances achieve comparable accuracies, with DTKP-AM  having a slight edge.
For SumN, DAMP provenance is more effective than the DTKP-AM by 72\% points on average,
since the top-k proofs cannot capture all the possible ways in which sums of digits can be computed.

In contrast, for HWF,  DTKP-AM  is more effective than DAMP by an average of 42.2\% points.
Each step of the HWF program, shown in Appendix~\ref{app:hwf}, involves both a concatenation operation and a partial parsing operation before the final expression is evaluated to produce a result. As such, it is difficult for the tags in DAMP to capture the semantics of the symbolic program.
In the case of DTKP-AM, each tag is a collection of proofs over input symbols corresponding to logits derived from the neural model. Therefore, any calculated gradients can be directly backpropagated to the logits that most influenced the output, making this a more effective provenance for this task.

\vspace{-0.05in}
\section{Discussion}
\label{sec:discussion}

Dolphin can be used for any task where the output of a model can be cast as a distribution over probabilities. This abstraction naturally encompasses a wide range of discriminative models in machine learning, such as classifiers, structured prediction systems, and vision-language models. By associating each symbolic object with a probability distribution over its possible values (tags), Dolphin enables downstream symbolic reasoning over uncertain predictions made by neural networks.

For instance, consider an autonomous driving scenario where a standard object detector such as Faster R-CNN outputs bounding boxes and class probabilities. One can represent each detection as a symbolic object whose attributes include coordinates and a \texttt{Distribution} over class probabilities:

\begin{lstlisting}[style=PythonStyle, frame=none, escapechar=@]
CLASSES = ['car', 'person', 'truck', ...]

class DetectedObject:
    def __init__(self, coords, score, class_logits):
        self.coords = coords
        self.score = score
        self.distr = Distribution(CLASSES, class_logits)
\end{lstlisting}
\vspace{-0.2in}
Now, consider checking whether a detected person is inside a detected car using symbolic reasoning:
\begin{lstlisting}[style=PythonStyle, frame=none, escapechar=@]
    def is_inside(coord_a, coord_b):
        ...

    person_inside_car = apply(
        o1.distr, o2.distr, lambda c1, c2: 
        c1 == "person" and c2 == "car" and 
        is_inside(o1.coords, o2.coords)
    )
\end{lstlisting}
\vspace{-0.1in}
This computation yields a \texttt{Distribution} over \texttt{True} and \texttt{False}, representing the likelihood that two detected objects constitute a person inside a car. This paradigm can generalize to many domains where neural predictions must be interpreted symbolically—for example, relationship inference in vision scenes, symbolic post-processing over structured outputs, or probabilistic parsing in NLP. Dolphin thus offers a principled and composable way to integrate symbolic logic over deep model outputs, opening the door for broader real-world applicability.

\section{Related Work}

\mypara{Neurosymbolic programming frameworks} Apart from existing frameworks like Scallop~\citep{li2023scallop}, DeepProbLog~\citep{manhaeve2018deepproblog}, and ISED~\citep{solko2024data}, there exist domain-specific tools such as NeurASP~\cite{yang2023neurasp} for phrase alignment. These approaches often suffer from inefficiencies due to CPU-bound symbolic reasoning.

\mypara{Scaling techniques} Various methods exist to scale differentiable reasoning. LYRICS~\citep{lyrics}, Logic Tensor Networks~\citep{ltn}, and Tensorlog~\citep{tensorlog} compile first-order logic constraints into GPU-compatible computation graphs. Other techniques, such as Greedy NTP~\citep{minervini2020differentiable} and the conditional theorem prover~\citep{minervini2020learning}, optimize proof search using heuristics. SLASH~\citep{slash} integrates neural networks and probabilistic circuits with Answer Set Programming, achieving scalability by dynamically pruning stochastically insignificant parts of programs. A-NESI~\citep{anesi} uses learned neural models to approximate the exact probabilistic semantics of WMC, boosting scalability. However, these methods are often task-specific and lack generalizability to broader neurosymbolic learning, or they struggle to scale due to memory constraints when grounding symbolic computations on GPU.

\mypara{Specialized neurosymbolic solutions}
There are many specialized solutions for various neurosymbolic tasks. For instance, NGS~\citep{li2020closed} uses a hand-coded syntax to specify the structure of mathematical expressions for HWF. More general solutions, such as NS-CL~\citep{mao2019neuro} includes a framework for visual question answering that learns symbolic representations for text and images. NeRd~\citep{chen2021neurallog} transforms questions in natural language into executable programs based on symbolic information extracted from text. \cite{seqrnn} proposes a recurrent neural network architecture that achieves 95\% accuracy on Path-32 and 94\% on Path-128. In contrast, \tool{} is a general system that tries to scale diverse neurosymbolic programs.

\section{Conclusion and Limitations}

We proposed \tool, a framework for scaling neurosymbolic learning.
\tool provides abstractions for writing symbolic programs along with pluggable vectorized provenances to compute symbolic gradients.
This allows users to write differentiable symbolic programs in Python within PyTorch pipelines that can scale
to complex programs and large datasets.
We show that \tool scales significantly better than existing neurosymbolic frameworks while achieving state-of-the-art performance on a variety of tasks.

A limitation of \tool is that it needs the user to write programs
in a batched manner. This is a common pattern within deep learning but may be restrictive to users new to batched programming.
Also, while \tool works well with most models, the representation needed
by generative models (e.g., Causal LLMs) has not been investigated yet. \tool{} also lacks support for non-deterministic symbolic programs. We leave these for future work.

\section*{Acknowledgements}
We thank the reviewers for their insightful feedback that helped to improve this paper. This research was supported by the ARPA-H program on Safe and Explainable AI under the award D24AC00253-00, the NSF award \#2313010, and a Google PhD Fellowship.

\section*{Impact Statement}

This paper presents work whose goal is to advance the field of Machine Learning. There are many potential societal consequences of our work, none which we feel must be specifically highlighted here.

\bibliography{references}

\begin{thebibliography}{30}
\providecommand{\natexlab}[1]{#1}
\providecommand{\url}[1]{\texttt{#1}}
\expandafter\ifx\csname urlstyle\endcsname\relax
  \providecommand{\doi}[1]{doi: #1}\else
  \providecommand{\doi}{doi: \begingroup \urlstyle{rm}\Url}\fi

\bibitem[Badreddine et~al.(2022)Badreddine, {d'Avila Garcez}, Serafini, and
  Spranger]{ltn}
Badreddine, S., {d'Avila Garcez}, A., Serafini, L., and Spranger, M.
\newblock Logic tensor networks.
\newblock \emph{Artificial Intelligence}, 303:\penalty0 103649, 2022.
\newblock ISSN 0004-3702.
\newblock \doi{https://doi.org/10.1016/j.artint.2021.103649}.
\newblock URL
  \url{https://www.sciencedirect.com/science/article/pii/S0004370221002009}.

\bibitem[Chaudhuri et~al.(2021)Chaudhuri, Ellis, Polozov, Singh, Solar-Lezama,
  Yue, et~al.]{chaudhuri2021neurosymbolic}
Chaudhuri, S., Ellis, K., Polozov, O., Singh, R., Solar-Lezama, A., Yue, Y.,
  et~al.
\newblock Neurosymbolic programming.
\newblock \emph{Foundations and Trends{\textregistered} in Programming
  Languages}, 7\penalty0 (3):\penalty0 158--243, 2021.

\bibitem[Chen et~al.(2021)Chen, Gao, and Moss]{chen2021neurallog}
Chen, Z., Gao, Q., and Moss, L.~S.
\newblock {N}eural{L}og: Natural language inference with joint neural and
  logical reasoning.
\newblock In Ku, L.-W., Nastase, V., and Vuli{\'c}, I. (eds.),
  \emph{Proceedings of *SEM 2021: The Tenth Joint Conference on Lexical and
  Computational Semantics}, pp.\  78--88, Online, August 2021. Association for
  Computational Linguistics.
\newblock \doi{10.18653/v1/2021.starsem-1.7}.
\newblock URL \url{https://aclanthology.org/2021.starsem-1.7}.

\bibitem[Cohen et~al.(2020)Cohen, Yang, and Mazaitis]{tensorlog}
Cohen, W.~W., Yang, F., and Mazaitis, K.
\newblock Tensorlog: A probabilistic database implemented using deep-learning
  infrastructure.
\newblock \emph{J. Artif. Intell. Res.}, 67:\penalty0 285--325, 2020.
\newblock URL \url{https://api.semanticscholar.org/CorpusID:211263674}.

\bibitem[Dang et~al.(2021)Dang, Khosravi, Liang, Vergari, and den
  Broeck]{juice}
Dang, M., Khosravi, P., Liang, Y., Vergari, A., and den Broeck, G.~V.
\newblock Juice: A julia package for logic and probabilistic circuits.
\newblock In \emph{AAAI Conference on Artificial Intelligence}, 2021.
\newblock URL \url{https://api.semanticscholar.org/CorpusID:235363700}.

\bibitem[Darwiche(2020)]{darwiche2020advance}
Darwiche, A.
\newblock An advance on variable elimination with applications to tensor-based
  computation.
\newblock In \emph{ECAI 2020}, pp.\  2559--2568. IOS Press, 2020.

\bibitem[De~Smet et~al.(2024)De~Smet, Sansone, and Zuidberg
  Dos~Martires]{de2024differentiable}
De~Smet, L., Sansone, E., and Zuidberg Dos~Martires, P.
\newblock Differentiable sampling of categorical distributions using the
  catlog-derivative trick.
\newblock \emph{Advances in Neural Information Processing Systems}, 36, 2024.

\bibitem[Green et~al.(2007)Green, Karvounarakis, and
  Tannen]{green2007provenance}
Green, T.~J., Karvounarakis, G., and Tannen, V.
\newblock Provenance semirings.
\newblock In \emph{Proceedings of the twenty-sixth ACM SIGMOD-SIGACT-SIGART
  symposium on Principles of database systems}, pp.\  31--40, 2007.

\bibitem[Hayes et~al.(2022)Hayes, Zhang, Yin, Pang, Sheng, Yang, Ge, Hu, and
  Parikh]{hayes2022Mugen}
Hayes, T., Zhang, S., Yin, X., Pang, G., Sheng, S., Yang, H., Ge, S., Hu, Q.,
  and Parikh, D.
\newblock Mugen: A playground for video-audio-text multimodal understanding and
  generation.
\newblock In \emph{European Conference on Computer Vision}, pp.\  431--449.
  Springer, 2022.

\bibitem[Huang et~al.(2021)Huang, Li, Chen, Samel, Naik, Song, and
  Si]{huang2021scallop}
Huang, J., Li, Z., Chen, B., Samel, K., Naik, M., Song, L., and Si, X.
\newblock Scallop: From probabilistic deductive databases to scalable
  differentiable reasoning.
\newblock \emph{Advances in Neural Information Processing Systems},
  34:\penalty0 25134--25145, 2021.

\bibitem[Kambhampati et~al.()Kambhampati, Valmeekam, Guan, Verma, Stechly,
  Bhambri, Saldyt, and Murthy]{kambhampati2024llms}
Kambhampati, S., Valmeekam, K., Guan, L., Verma, M., Stechly, K., Bhambri, S.,
  Saldyt, L.~P., and Murthy, A.~B.
\newblock Position: Llms can’t plan, but can help planning in llm-modulo
  frameworks.
\newblock In \emph{Forty-first International Conference on Machine Learning}.

\bibitem[Li et~al.(2020)Li, Huang, Hong, Chen, Wu, and Zhu]{li2020closed}
Li, Q., Huang, S., Hong, Y., Chen, Y., Wu, Y.~N., and Zhu, S.-C.
\newblock Closed loop neural-symbolic learning via integrating neural
  perception, grammar parsing, and symbolic reasoning.
\newblock In \emph{International Conference on Machine Learning}, pp.\
  5884--5894. PMLR, 2020.

\bibitem[Li et~al.(2023)Li, Huang, and Naik]{li2023scallop}
Li, Z., Huang, J., and Naik, M.
\newblock Scallop: A language for neurosymbolic programming.
\newblock \emph{Proceedings of the ACM on Programming Languages}, 7\penalty0
  (PLDI):\penalty0 1463--1487, 2023.

\bibitem[Liu(2019)]{liu2019roberta}
Liu, Y.
\newblock Roberta: A robustly optimized bert pretraining approach.
\newblock \emph{arXiv preprint arXiv:1907.11692}, 2019.

\bibitem[Manhaeve et~al.(2018)Manhaeve, Dumancic, Kimmig, Demeester, and
  De~Raedt]{manhaeve2018deepproblog}
Manhaeve, R., Dumancic, S., Kimmig, A., Demeester, T., and De~Raedt, L.
\newblock Deepproblog: Neural probabilistic logic programming.
\newblock \emph{Advances in neural information processing systems}, 31, 2018.

\bibitem[Mao et~al.(2019)Mao, Gan, Kohli, Tenenbaum, and Wu]{mao2019neuro}
Mao, J., Gan, C., Kohli, P., Tenenbaum, J.~B., and Wu, J.
\newblock The neuro-symbolic concept learner: Interpreting scenes, words, and
  sentences from natural supervision.
\newblock In \emph{International Conference on Learning Representations}, 2019.
\newblock URL \url{https://openreview.net/forum?id=rJgMlhRctm}.

\bibitem[Marra et~al.(2019)Marra, Giannini, Diligenti, and Gori]{lyrics}
Marra, G., Giannini, F., Diligenti, M., and Gori, M.
\newblock Lyrics: A general interface layer to integrate logic inference and
  deep learning.
\newblock In \emph{Machine Learning and Knowledge Discovery in Databases:
  European Conference, ECML PKDD 2019, W\"{u}rzburg, Germany, September
  16–20, 2019, Proceedings, Part II}, pp.\  283–298, Berlin, Heidelberg,
  2019. Springer-Verlag.
\newblock ISBN 978-3-030-46146-1.
\newblock \doi{10.1007/978-3-030-46147-8_17}.
\newblock URL \url{https://doi.org/10.1007/978-3-030-46147-8_17}.

\bibitem[Minervini et~al.(2020{\natexlab{a}})Minervini, Bo{\v{s}}njak,
  Rockt{\"a}schel, Riedel, and Grefenstette]{minervini2020differentiable}
Minervini, P., Bo{\v{s}}njak, M., Rockt{\"a}schel, T., Riedel, S., and
  Grefenstette, E.
\newblock Differentiable reasoning on large knowledge bases and natural
  language.
\newblock In \emph{Proceedings of the AAAI conference on artificial
  intelligence}, volume~34, pp.\  5182--5190, 2020{\natexlab{a}}.

\bibitem[Minervini et~al.(2020{\natexlab{b}})Minervini, Riedel, Stenetorp,
  Grefenstette, and Rockt{\"a}schel]{minervini2020learning}
Minervini, P., Riedel, S., Stenetorp, P., Grefenstette, E., and
  Rockt{\"a}schel, T.
\newblock Learning reasoning strategies in end-to-end differentiable proving.
\newblock In \emph{International Conference on Machine Learning}, pp.\
  6938--6949. PMLR, 2020{\natexlab{b}}.

\bibitem[Naik et~al.(2024)Naik, Stein, Wu, Naik, and Wong]{torchql}
Naik, A., Stein, A., Wu, Y., Naik, M., and Wong, E.
\newblock Torchql: A programming framework for integrity constraints in machine
  learning.
\newblock \emph{Proc. ACM Program. Lang.}, 8\penalty0 (OOPSLA1), April 2024.
\newblock \doi{10.1145/3649841}.
\newblock URL \url{https://doi.org/10.1145/3649841}.

\bibitem[Orvieto et~al.(2023)Orvieto, Smith, Gu, Fernando, Gulcehre, Pascanu,
  and De]{seqrnn}
Orvieto, A., Smith, S.~L., Gu, A., Fernando, A., Gulcehre, C., Pascanu, R., and
  De, S.
\newblock Resurrecting recurrent neural networks for long sequences.
\newblock In \emph{Proceedings of the 40th International Conference on Machine
  Learning}, ICML'23. JMLR.org, 2023.

\bibitem[Sanh(2019)]{sanh2019distilbert}
Sanh, V.
\newblock Distilbert, a distilled version of bert: Smaller, faster, cheaper and
  lighter.
\newblock \emph{arXiv preprint arXiv:1910.01108}, 2019.

\bibitem[{Scallop Language Group}(2022)]{ssft2022provenance}
{Scallop Language Group}.
\newblock Scallop and neuro-symbolic programming: Tags, instrumentation, and
  provenance.
\newblock Eleventh Summer School on Formal Techniques, 2022.
\newblock URL \url{https://www.scallop-lang.org/ssft22/lectures/lecture-2.pdf}.

\bibitem[Serafini \& Garcez(2016)Serafini and Garcez]{serafini2016logic}
Serafini, L. and Garcez, A.~d.
\newblock Logic tensor networks: Deep learning and logical reasoning from data
  and knowledge.
\newblock \emph{arXiv preprint arXiv:1606.04422}, 2016.

\bibitem[Sinha et~al.(2019)Sinha, Sodhani, Dong, Pineau, and
  Hamilton]{sinha2019clutrr}
Sinha, K., Sodhani, S., Dong, J., Pineau, J., and Hamilton, W.~L.
\newblock {CLUTRR}: A diagnostic benchmark for inductive reasoning from text.
\newblock In Inui, K., Jiang, J., Ng, V., and Wan, X. (eds.), \emph{Proceedings
  of the 2019 Conference on Empirical Methods in Natural Language Processing
  and the 9th International Joint Conference on Natural Language Processing
  (EMNLP-IJCNLP)}, pp.\  4506--4515, Hong Kong, China, November 2019.
  Association for Computational Linguistics.
\newblock \doi{10.18653/v1/D19-1458}.
\newblock URL \url{https://aclanthology.org/D19-1458}.

\bibitem[Skryagin et~al.(2024)Skryagin, Ochs, Dhami, and Kersting]{slash}
Skryagin, A., Ochs, D., Dhami, D.~S., and Kersting, K.
\newblock Scalable neural-probabilistic answer set programming.
\newblock \emph{J. Artif. Int. Res.}, 78, January 2024.
\newblock ISSN 1076-9757.
\newblock \doi{10.1613/jair.1.15027}.
\newblock URL \url{https://doi.org/10.1613/jair.1.15027}.

\bibitem[Solko-Breslin et~al.(2024)Solko-Breslin, Choi, Li, Velingker, Alur,
  Naik, and Wong]{solko2024data}
Solko-Breslin, A., Choi, S., Li, Z., Velingker, N., Alur, R., Naik, M., and
  Wong, E.
\newblock Data-efficient learning with neural programs.
\newblock \emph{arXiv preprint arXiv:2406.06246}, 2024.

\bibitem[Tay et~al.(2021)Tay, Dehghani, Abnar, Shen, Bahri, Pham, Rao, Yang,
  Ruder, and Metzler]{tay2020long}
Tay, Y., Dehghani, M., Abnar, S., Shen, Y., Bahri, D., Pham, P., Rao, J., Yang,
  L., Ruder, S., and Metzler, D.
\newblock Long range arena : A benchmark for efficient transformers.
\newblock In \emph{International Conference on Learning Representations}, 2021.
\newblock URL \url{https://openreview.net/forum?id=qVyeW-grC2k}.

\bibitem[van Krieken et~al.(2023)van Krieken, Thanapalasingam, Tomczak, van
  Harmelen, and Ten~Teije]{anesi}
van Krieken, E., Thanapalasingam, T., Tomczak, J., van Harmelen, F., and
  Ten~Teije, A.
\newblock A-nesi: A scalable approximate method for probabilistic neurosymbolic
  inference.
\newblock In Oh, A., Naumann, T., Globerson, A., Saenko, K., Hardt, M., and
  Levine, S. (eds.), \emph{Advances in Neural Information Processing Systems},
  volume~36, pp.\  24586--24609. Curran Associates, Inc., 2023.

\bibitem[Yang et~al.(2021)Yang, Ishay, and Lee]{yang2023neurasp}
Yang, Z., Ishay, A., and Lee, J.
\newblock Neurasp: embracing neural networks into answer set programming.
\newblock In \emph{Proceedings of the Twenty-Ninth International Joint
  Conference on Artificial Intelligence}, IJCAI'20, 2021.
\newblock ISBN 9780999241165.

\end{thebibliography}
\bibliographystyle{icml2025}

\newpage
\appendix
\onecolumn

\section{DTKP-AM Provenance}
\label{app:dtkp}
\changed{We clarify and expand on some aspects of the DTKP-AM provenance.}

\changed{\subsection{WMC approximation}}
\changed{In this section, we emphasize that DTKP-AM does not perform precise weighted model counting (WMC) and address possible shortcomings that could arise.
A hardware-efficient vectorization of exact WMC is beyond the scope of this paper, and is itself an active area of research.
Instead, we use the following add-mult approximation of WMC:
\[
\Pr(t) = \sum_i \Pr(t_i) = \sum_i \prod_j \text{norm}(t_{ij})
\]
We note that this approximation upper bounds the result from DTKP-WMC: the coarseness arises from the summation, which may double count models that satisfy more than one of the proofs.
However, add-mult achieves significant computational speedup since it simplifies the exponential enumeration over all possible models into a linear pass over the tag's elements.}

\changed{We further claim that this approximation does not destroy \textit{all} the semantics from DTKP-WMC due to DTKP-AM’s faithful implementation of the semiring operations $\oplus$ and $\otimes$ for tracking top-k proofs.
DTKP-AM tags therefore remain similar to DTKP-WMC tags at every intermediate symbolic reasoning step.
By contrast, the imprecise add-mult is a one-time transformation of the final tags into probabilities, performed only after the tags have been propagated through the entire symbolic program.
Crucially, we show there exists information that is uniquely captured by top-k tag operations, and is not lost when fuzzily converting the tags to probabilities.}

\changed{As a simple illustrative example, consider using \textsc{Apply} with the following toy function:
\begin{align*}
    f(a, b) = \begin{cases} \mathbb T & a = b \\ \mathbb F & \text{otherwise} \end{cases}
\end{align*}
For any distribution $D$ of mutually exclusive input symbols (e.g. the digit classification of a CNN), we intuitively would like the distribution $f(D, D)$ to assign a probability of 1 to symbol $\mathbb T$ and a probability of 0 to symbol $\mathbb F$.
According to our semantics, the tag for $\mathbb T$ is actually given by:
\[
    f(D, D)(\mathbb T) = \bigoplus_i (D(i)\otimes D(i))
\]
However, if we were to use DAMP to compute the tags for $f(D, D)$, the provenance treats the two input distributions as independent when they are the exact same distribution! Thus, the probability assigned to $\mathbb T$ by $f(D, D)$ is incorrectly calculated as:
\[
    f(D, D)(\mathbb T) = \sum_i (D(i))^2
\]
On the other hand, consider any top-k provenance that satisfies:
\[
    t \otimes t' = \text{top}_k(\{ t_i \cup t'_j \mid (t_i, t'_j)\in t\times t' \})
\]
where $\times$ is the set Cartesian product.
Note that DTKP-AM does satisfy this condition, where the set union is implemented with an element-wise minimum.
Now assuming $D(i)$ is initialized in the natural way (i.e. a tag consisting of a single proof containing just the input symbol $i$), then $D(i)\otimes D(i)=D(i)$ and therefore:
\[
    f(D, D)(\mathbb T) = \bigoplus_i D(i)
\]
Under both DTKP-WMC and DTKP-AM, the probability of $\mathbb T$ is:
\[
    \sum_i \prod_j \text{norm}(D(i)_{ij}) = \sum_i D(i)_{ii} = \sum_i \Pr(i) = 1
\]
for any normalized $D$ with at most $k$ symbols.
Even if the number of symbols exceeds $k$, we note that the distributions we seek to learn are often skewed (an accurate model should assign a probability to the ground truth that significantly outweighs the other symbols).
For such distributions, DTKP and DTKP-AM would still yield the same probability for $\mathbb T$, and it is much closer to 1 than the sum of squares result from DAMP.}

\changed{While this example may seem contrived, it still suggests the smaller role a ``correct" WMC can have on the final answer compared to $\oplus$ and $\otimes$ implemented with proper set-based semantics.
We even hypothesize that in most cases, the add-mult approximation does not meaningfully affect the final result compared to DTKP-WMC.
This is empirically demonstrated by our benchmark results, which shows DTKP-AM achieving similar accuracy to Scallop's implementation of DTKP-WMC.
In fact, DAMP can be considered as a sort of ablation, where both the WMC and semiring operations use fuzzy add-mult semantics instead of a set-based one, and indeed, its accuracy often performs worse than both DTKP-WMC and DTKP-AM.}

\changed{\subsection{Role of $+\infty$ and $-\infty$}}
\changed{In this section, we motivate the use of $+\infty$ and $-\infty$ in DTKP-AM's tensor representation of tags.
Because tensors $t$ are rectangular where every proof $i$ and symbol $j$ must have an entry $t_{ij}$, we require a way to denote the absence of an input symbol from a proof, and the absence of a proof from a tag.
Importantly, an absent symbol should not influence the probability of a proof (i.e. its normalized value should contribute 1 to the probability's product), and an absent proof should not influence the probability of a tag (i.e. it should contribute 0 to the sum during add-mult WMC).
Indeed this is captured by our definition of $\text{norm}$, which clamps $+\infty$ to 1  (representing absent symbols) and $-\infty$ to 0 (representing absent proofs) during any probability calculation. While this introduces clamping operations, PyTorch's implementation of clamp backpropagation ensures a gradient of 1 everywhere, even on the clamp boundaries (source: \url{https://github.com/pytorch/pytorch/pull/7049}).}

\changed{Since $\mathbf{\hat 0}$ corresponds to the tag consisting of no proofs (i.e. a tag with probability 0), we initialize it to be a tensor where every proof is absent (all $-\infty$).
Likewise, since $\mathbf{\hat 1}$ corresponds to the tag consisting of a single empty proof (i.e. a tag with probability 1), we initialize it to be a tensor where every symbol is absent from the first row / proof (all $+\infty$), while the remaining rows / proofs are absent (all $-\infty$).}

\changed{\subsection{Further reading}}
\changed{For a more in-depth explanation of provenances in general, including the formalization of DTKP semantics with Boolean formulae, see Section 4 of \cite{li2023scallop}. 
For worked examples of provenance computation with comparisons of top-k provenances to DAMP, we refer the reader to \cite{ssft2022provenance}.}

\section{Control Flows and Recursion in \tool{}}
\label{app:control_flow}

\begin{table}[h]
\footnotesize
\centering
\caption{\changed{Time taken by the symbolic program for the HWF task split by the time spent on the CPU and GPU. UDFs refer to user-defined functions where control flows reside for HWF. The times annotated with C and G indicate time spent on the CPU and GPU, respectively.}}
\begin{tabular}{lccc}
\toprule
\textbf{Config} & \textbf{Time for UDF (s)} & \textbf{Time for Tag Computations (s)} & \textbf{Total Time (s)} \\
\midrule
No Parallelism & 36.24 (C) & 461.02 (C) & 497.26 \\
Parallelized Tag Computations & 14.13 (C) & 75.125 (G) & 89.25 \\
\bottomrule
\end{tabular}
\label{tab:control_flow}
\end{table}

\begin{figure}[h]
\centering
\begin{lstlisting}[style=PythonStyle, frame=none, escapechar=@, basicstyle=\ttfamily\footnotesize]
Coord = namedtuple('Coord', ['x', 'y'])
    
def compute_paths(paths, edges):
    new_paths = apply_if(paths, edges, lambda p1, p2: Coord(p1.x, p2.y), lambda p1, p2: p1.y == p2.x)
    merged = union(paths, new_paths)
    # checking for convergence via fix-point
    if merged.symbols == paths.symbols:
        return merged
    else:
        return compute_paths(merged, edges)

edges = Distribution(model(img), points)
paths = compute_paths(edges, edges)
\end{lstlisting}
\caption{\changed{Example of a transitive closure computation in \tool{}.}}
\label{fig:transitive_closure_app}
\end{figure}

\changed{In this section, we provide a more detailed explanation of how \tool{} handles control flows and recursion.
In \tool{}, control flows largely exist within the lambda functions supplied to the `Apply`, `ApplyIf`, and `Filter` operations, which can be arbitrary Python functions over the symbols in the Distributions. As discussed in Section~\ref{sec:operations}, these functions can include complex operations like if-then-else branches, loops, and even recursions.
We do assume that divergent control flows are resolved within the lambda function itself. The nature of these functions means that they cannot be parallelized over the GPU. Instead, they are executed sequentially on the CPU, while the associated tags are computed parallely on the GPU. We optimize the design of the Distribution class so that there is one set of CPU-based computations for the entire batch of samples rather than one set of computations for each sample, which is typical of other neurosymbolic frameworks. This allows \tool{} to maintain the benefits of parallelism even while the user-defined functions are executed sequentially.}

\changed{\subsection{Control Flow in HWF}}

\changed{We demonstrate this by showing the time taken by the symbolic program for the HWF task split by the time spent on the CPU and GPU in Table~\ref{tab:control_flow}. The first row shows the time taken when the neurosymbolic model is run sequentially on the CPU with no parallelism. The second row shows the time taken when tag computations are parallelized on the GPU over batches of 64 samples each. The times annotated with C and G indicate time spent on the CPU and GPU, respectively. We only show the time taken in the forward pass in the table.}

\changed{Observe that the time, both for UDF computation and for Tag computation, decreases as we move from sequential CPU evaluation to the batched evaluation. Due to \tool{}’s design, increases in batch size result in fewer CPU operations, since the set of CPU operations is shared for the entire batch, while parallelizing more tag computations over the entire batch.}

\changed{\subsection{Recursion}}

\changed{In order to write recursive computations in \tool{}, one has two choices: either supply a recursive user-defined function to the \tool{} primitives, or write a more fine-grained program in Python that uses \tool{} primitives in the base case as well as the recursive case, set to terminate once a condition is met. Here, the diverging control flows can be merged using the \textsc{Union} primitive. We follow the latter approach for tasks involving recursion, such as Path and CLUTRR. The crux of those programs involves performing a transitive closure computation over a graph, represented by a set of edges for Path or relations for CLUTRR. We show an example of a transitive closure computation in Figure~\ref{fig:transitive_closure_app}.}

\changed{Here, lets say that \code{model} is a neural model that predicts the edges between each pair of points in a graph, represented by \code{points}. The \code{compute\_paths} function computes the transitive closure of the graph by iteratively applying the edges to the paths. The \textsc{ApplyIf} function applies the edges to the paths if the end of the first path is the same as the start of the second path. The \textsc{Union} function merges the new paths with the existing paths. The function \code{compute\_paths} is called recursively until a fixpoint is reached, specifically until no new paths can be added. This is a simple example of a recursive computation in \tool{}, and also forms the core program needed for the PathFinder task. We perform a similar recursive computation for the CLUTRR task, where we find the transitive closure of a graph representing relations between people in a passage.}

\section{On the Language and Semantics}
\changed{\subsection{Language}}
\changed{To develop the operations provided by \tool{}, we studied several neurosymbolic tasks to determine the most common operations needed for these tasks. We found that the main operation needed in most programs is to apply a function to symbols from different input models and relations. This is primarily achieved via the join operation in Datalog, but we introduce the \code{Apply} or \code{ApplyIf} primitives for a more Pythonic approach. \code{Filter}s are used to remove symbols violating conditions, similar to Datalog selections, while \code{Union} mimics the disjunction operation in Datalog, typically needed for writing recursive programs as described in Appendix~\ref{app:control_flow}.
}

\changed{\subsection{Semantics}}
\changed{We designed \tool{} to be a general-purpose neurosymbolic framework able to support various semantics, as long as they can be expressed as operations over tags tracked via the Distribution class. \tool{} assumes that the provenance supplied to it offers both the conjunction and disjunction operations that operate over combinations of tags from input symbols, as well as a way to translate tags to probabilities. As long as these assumptions are satisfied, the primitives of \tool{} preserve the semantics offered by the provenances.}

\changed{As such, supplying the DAMP provenance to the \tool{} program introduces basic fuzzy semantics which are preserved by \tool{}. However, there are cases where the independence assumptions may not hold and fuzzy semantics may not be appropriate.}

\changed{The DTKP-AM provenance, on the other hand, offers an alternative without the assumption of variable independence, except on the input variables.
At each step of the program, each symbol is associated with the tags of the input symbols that produce it via the proofs. Again, since DTKP-AM satisfies the aforementioned assumptions, the top-k semantics of the provenance are preserved.}

\changed{These tags are then translated into probabilities by performing an add-mult operation over the proofs. This approximation of the WMC operation is more complex and results in a more precise translation of tags to probabilities. However, as we see in the experiments where Scallop uses DTKP-WMC, the accuracies achieved by DTKP-AM and DTKP-WMC are comparable.}

\section{\tool Experiment Details for Benchmarks}
\label{app:code}

\subsection{Comparison of Per Epoch Training Times}
\label{app:training_time_table}
\begin{table*}[h]
    \setlength{\tabcolsep}{0.5em}
    \scriptsize
    \centering
    \caption{Comparison of training times \cameraready{(in seconds)} taken by each baseline. The Timeout (TO) is set at 10 hours. $\alpha$ is the scaling factor, which is the ratio of the per epoch training times of the baselines and \tool.}
    \begin{tabular}{l r | rr | rr | rr | rr }
    \toprule
    \textbf{Task} & \multicolumn{1}{c|}{\textbf{\tool}} & \multicolumn{2}{c|}{\textbf{Scallop}} & \multicolumn{2}{c|}{\textbf{LTN}} & \multicolumn{2}{c|}{\textbf{ISED}} & \multicolumn{2}{c}{\textbf{IndeCateR+}} \\ 
    \midrule
    &  
      $\text{T}_{\text{epoch}}$ &
      $\text{T}_{\text{epoch}}$ & $\textbf{$\alpha$}_{\text{epoch}}$ &
      $\text{T}_{\text{epoch}}$ & $\textbf{$\alpha$}_{\text{epoch}}$ & 
      $\text{T}_{\text{epoch}}$ & $\textbf{$\alpha$}_{\text{epoch}}$ & 
      $\text{T}_{\text{epoch}}$ & $\textbf{$\alpha$}_{\text{epoch}}$\\
    \midrule
    SumN-5  & 10.77  
             &  184.76 & 17.16
             & 4.63
             & 0.43 & 59.93 & 5.56 & 59.54 & 5.52 \\
    SumN-10 & 10.49
             & 341.57 & 32.56 
             & \textbf{\textcolor{red}{OOM}} & --
              & 216.54 & 20.64  & 32.14  & 3.06 \\
    SumN-15 & 10.47
             & 493.87  & 47.17
             & \textbf{\textcolor{red}{OOM}}  & --
              & 653.39 & 62.41 & 23.84 & 2.28 \\
    \hline
    HWF-7 & 152.87
            & 499.57 & 3.27
            &  \multicolumn{2}{c|}{\multirow{3}{*}{N.A.}}
            &  201.09 & 1.32 & 540.26 & 3.53 \\
    HWF-15 & 858.8 &  \scnote{1.49}{4} & 17.35
            &  & 
             & \scnote{1.16}{3} & 1.35 & \scnote{2.51}{3} & 2.93 \\
    HWF-19 & \scnote{1.4}{3} & \scnote{3.92}{5} & 280 
            &  & 
            & \scnote{1.05}{4} & 7.5 & \scnote{4.18}{3} & 2.99  \\
    \midrule
    Path-32 & \scnote{1.29}{3}
              & \scnote{2.2}{3} & 1.71 
             & \multicolumn{6}{c}{\multirow{3}{*}{N.A.}}\\
    Path-128 & \scnote{1.67}{3} & \scnote{4.18}{3} & 2.5 
              \\
    Path-256 & \scnote{1.97}{3}
             & \scnote{1.13}{4} &  5.74
             & \\
    \midrule
    CLUTRR-3 & 152.21
               & 429.97  & 2.82
               & \multicolumn{6}{c}{\multirow{2}{*}{N.A.}} \\
    CLUTRR-4 & 165.13 &  783.11  & 4.74
               & \\
    \midrule
    Mugen-1K & 165.74
              & 133.68  & 0.81 
              & \multicolumn{6}{c}{\multirow{2}{*}{N.A.}} \\
    Mugen-5K & 826.31
              & 634.86 & 0.77
              & \\
    \bottomrule
\end{tabular}

    \label{tab:scalability}
    
\end{table*}

The following are more details on the experiment setup.
For each experimental trial, we report the highest evaluation accuracy over all epochs.
Unless otherwise noted, each trial was run on a machine with Intel Xeon Gold 6248 (2.50 GHz) CPUs and NVIDIA GeForce RTX 2080 Ti (11 GB) GPUs. Additionally, for Scallop, the accuracy of the best-performing provenance is reported. Table 5 shows the training time per epoch for all of the baselines in each of the benchmarks.

\subsection{Accuracy without Timeouts}
\label{app:acc_no_timeout}
\begin{figure}
    \centering
    \includegraphics[width=0.5\linewidth]{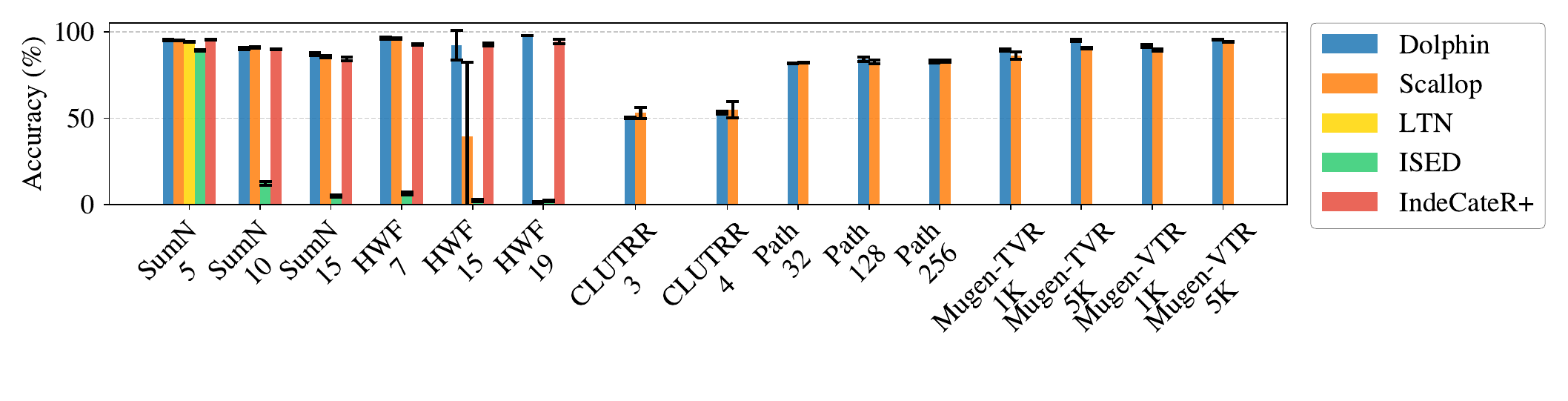}
    \caption{Accuracy of \tool{} and baselines across all benchmarks without any timeout.}
    \label{fig:acc_no_timeout}
\end{figure}

While Figure~\ref{fig:accuracy} shows the accuracy of baselines run until a timeout of 10 hours, Figure~\ref{fig:acc_no_timeout} shows their accuracy when run until convergence or until the test-time accuracy plateaus. We see that only IndeCateR+ for HWF-19 and Scallop for Path-256 are able to match \tool's accuracies. Scallop and ISED are unable to converge for the larger versions of HWF, while ISED is unable to converge for the larger versions of SumN.

\subsection{Effect of \texorpdfstring{$K$}{K} on Accuracy and Runtime}
\label{app:k_ablation}

We report preliminary results on the effect of the top-$K$ value in DTKP-AM across different tasks. As shown in Tables~\ref{tab:k_ablation_acc} and~\ref{tab:k_ablation_time}, increasing $K$ generally has little effect on final accuracy or per-epoch training time. HWF-19 is the only task where increasing $K$ offers noticeable gains. For all other benchmarks, accuracy remains stable and runtime scales sub-linearly due to DTKP-AM’s vectorized implementation.

\begin{table}[H]
\centering
\scriptsize
\caption{Accuracy (\%) of DTKP-AM with different values of $K$ across five benchmarks.}
\begin{tabular}{lcccc}
\toprule
\textbf{Benchmark} & \textbf{$K=1$} & \textbf{$K=3$} & \textbf{$K=5$} & \textbf{$K=7$} \\
\midrule
Sum-15    & 9.61         & 10.81        & 10.51        & 10.21 \\
HWF-19    & 8.94         & 99.15        & 96.89        & 95.75 \\
Path-256  & 81.39        & 82.14        & 80.86        & 82.38 \\
CLUTRR-4  & 53.62        & 48.52        & 50.35        & 48.17 \\
Mugen-5K  & (94.1 / 95.7)  & (95.4 / 95.7) & (95.3 / 95.4) & (95.4 / 95.4) \\
\bottomrule
\end{tabular}
\label{tab:k_ablation_acc}
\end{table}

\begin{table}[H]
\centering
\scriptsize
\caption{Training time per epoch (T/ep in seconds) for DTKP-AM with different values of $K$.}
\begin{tabular}{lcccc}
\toprule
\textbf{Benchmark} & \textbf{$K=1$} & \textbf{$K=3$} & \textbf{$K=5$} & \textbf{$K=7$} \\
\midrule
Sum-15    & 37.21   & 47.70   & 53.52   & 58.54 \\
HWF-19    & 1.21e3  & 1.40e3  & 1.33e3  & 1.46e3 \\
Path-256  & 1.97e3  & 2.34e3  & 2.10e3  & 2.12e3 \\
CLUTRR-4  & 240.50  & 257.89  & 261.31  & 257.99 \\
Mugen-5K  & 460.38  & 464.68  & 470.05  & 465.10 \\
\bottomrule
\end{tabular}
\label{tab:k_ablation_time}
\end{table}

\subsection{MNIST Sum-N}
\changed{For this task, the base neural network model is a standard CNN (a 3-layer convolutional network with ReLU activations) classifying each image into 10 classes of digits (0, 1, $\ldots$, 9). The symbolic module sums the Distribution objects over the logits output by the neural model for each image.}

Each of the MNIST Sum-N tasks had a batch train size of 64 samples, a learning rate of 0.001, and a top-k value of 1. Each of the tasks were trained on a dataset size of the original MNIST dataset divided by N of Sum-N. Sum5's dataset consisted of 12000 train samples and 2000 test samples. Sum10's dataset consisted of 6000 train samples and 1000 test samples. Sum15's dataset consisted of 4000 train samples and 666 test samples. Figure 7 is a high-level overview of the Sum-N model's architecture.

\begin{figure}[!h]
    \centering        
    \includegraphics[width=0.5\linewidth]{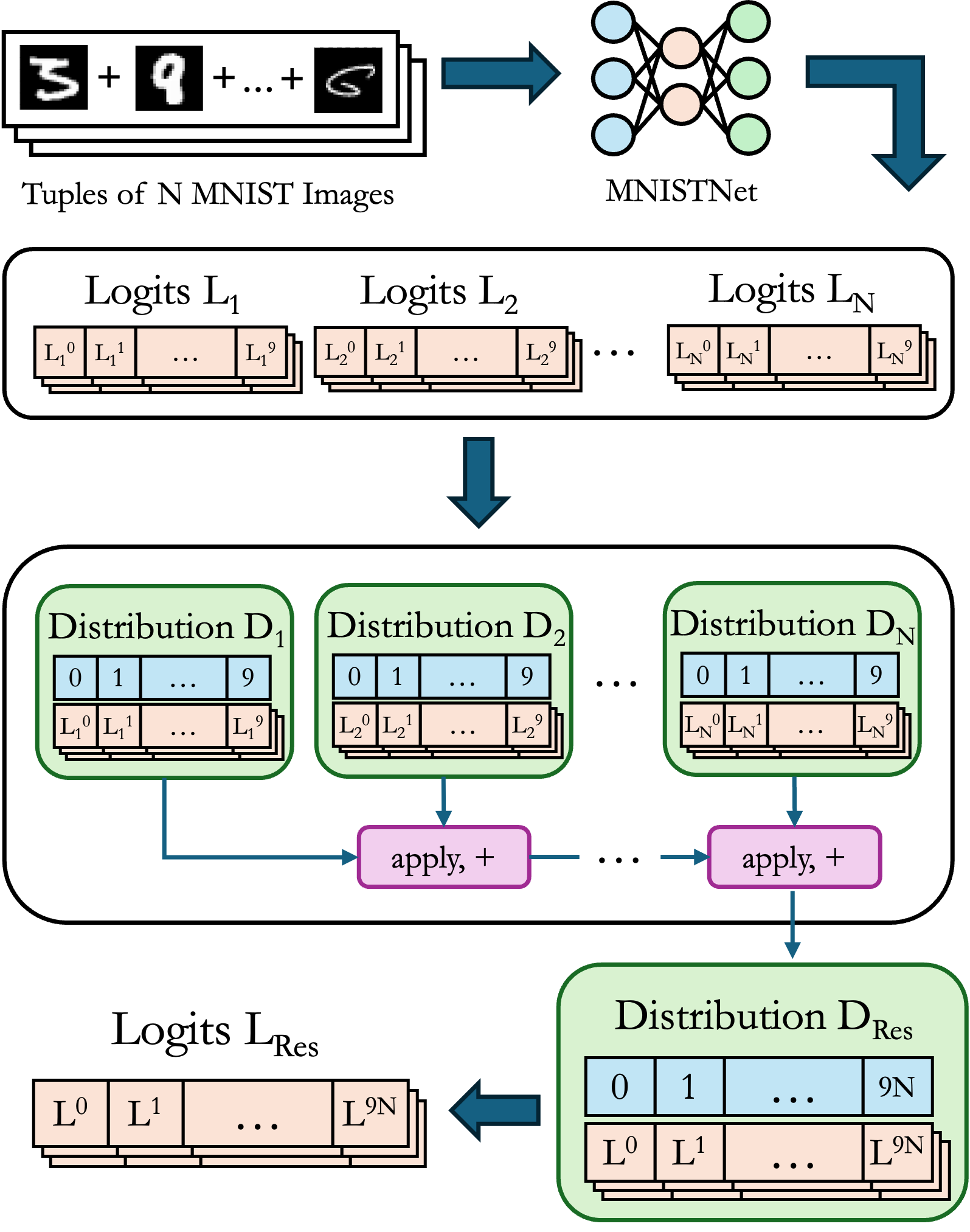}
    \caption{Components of the \code{SumNNet} model written in Figure~\ref{fig:mnist_sum_n_code}.}
    \label{fig:mnist_sum_breakdown}
\end{figure}

\subsection{Hand-Written Formula}
For Hand-Written Formula, \changed{the perception model is again a standard CNN that classifies images into 14 classes: 10 digits (0, 1, $\ldots$, 9), and 4 operations (+, -, $\times$, and /). The \tool program for this task builds strings of formulae from the outputs of the neural model and evaluates them
using Python's \texttt{eval} function, demonstrating the ability of \tool{} to support black-box functions.} 

We trained each task with a batch train size of 64 samples. The learning rate was 0.0001, the global sampling value was 7, and top-k value was 3. Length 7's dataset consisted of 9600 samples for training, 2400 samples for testing. Length 15 consisted of 24000 training samples and 6000 testing samples. Length 19 consisted of 32000 training samples and 8000 testing samples.

\subsection{PathFinder}
\changed{For this task, the perception model is also a CNN, but it predicts edges between pairs of nodes (denoted by dashes) as well as the end points depicted in the image of the maze. The \tool program for this task is recursive since it must search for paths between the two dots.}

For each of the PathFinder tasks, we used a batch train size of 64 samples, a learning rate of 0.0001, and a top-k value of 1. Each task's dataset consisted of 539459 images for training and 59940 images for testing. Each task had its own dataset of images with dimensions of the task's pixel size. 

\subsection{CLUTRR}
For each CLUTRR task, we used a single A100 GPU (40 GB), with a learning rate of 0.00001 and use a batch size of 16.
The length of the training dataset for CLUTRR (Small) was 11,093 and that of the test set was 1146.
The training set for CLUTRR (Medium) contained 15,083 samples and the test set contained 1048 samples.

\changed{The \tool{} program for CLUTRR receives as inputs pairs of entities from the input paragraph along with the logits for each pair over 21 possible relations produced by the classification head of the Roberta-base~\cite {liu2019roberta} model. The program then recursively derives relations over the graphs these pairs represent until no new relations can be derived. After that, it returns the Distribution over relations for the target pair of entities.}

\subsection{Mugen}
For each Mugen task, we use a batch size of 3 and a learning rate of 0.0001. From the full Mugen dataset, we sample a training set of 5000 examples for Mugen (Medium), and from that set, we sample a training set of 1000 for Mugen (Small). Both Small and Medium are evaluated on a fixed holdout set of 1000 samples.
We train and evaluate for up to 100 epochs.

\changed{
We use a combination of DistilBert~\citep{sanh2019distilbert} and S3D~\citep{de2024differentiable} as the perception model for the text and video inputs respectively.
The \tool{} program for both Mugen tasks computes the temporal alignment of a given text-video pair.
The inputs (extracted from the text and video by neural components) are pairs of IDs and actions, where the ID order corresponds to the action sequence (e.g. the IDs for video actions are the frame numbers).
The program finds the Distribution of all valid mappings between text event IDs and video frame IDs that preserve the order of actions.
}

\section{Graph Results of RQ3}
\label{app:prov_compare}
\begin{figure}[!h]
\begin{subfigure}{0.99\textwidth}
    \centering
    \includegraphics[width=0.85\linewidth]{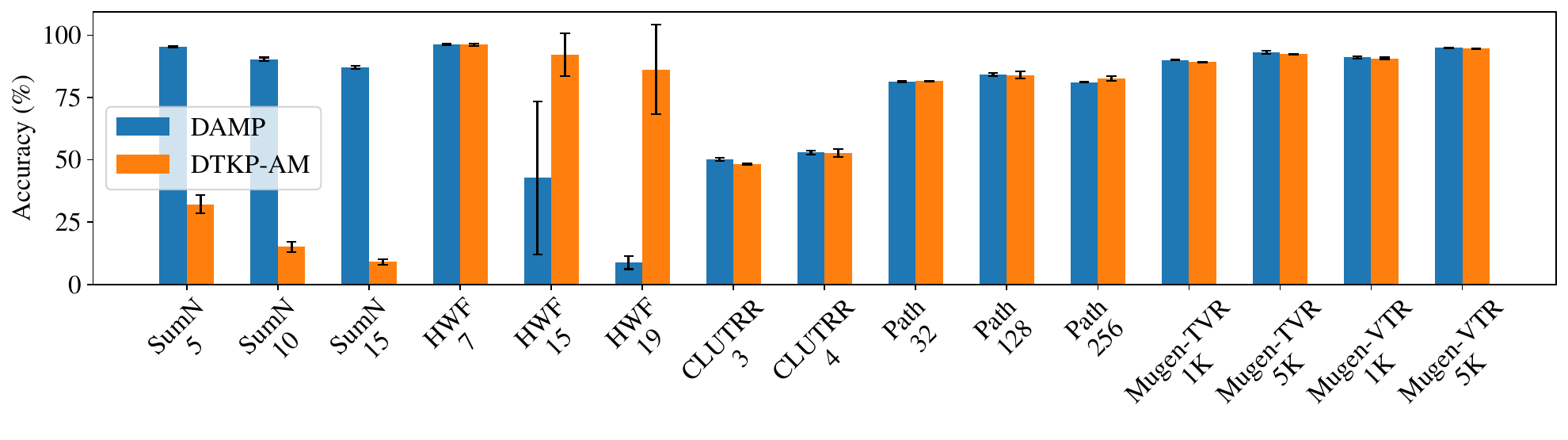}
\end{subfigure}
\begin{subfigure}{0.99\textwidth}
    \centering
    \includegraphics[width=0.85\linewidth]{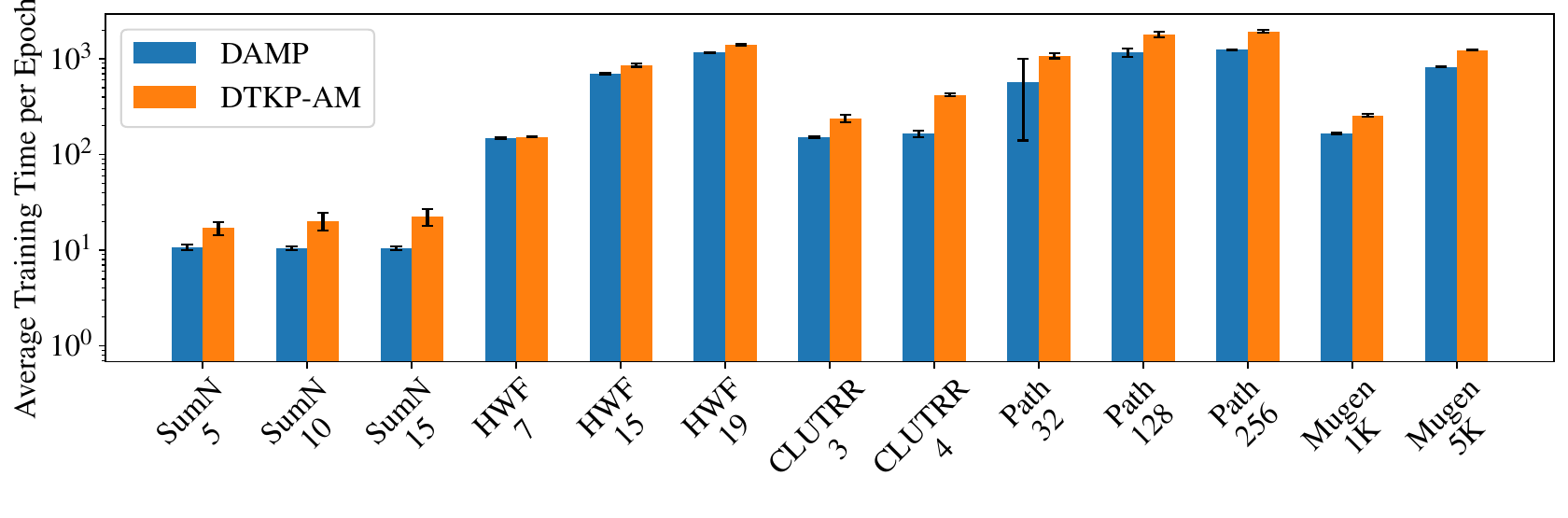}
    \end{subfigure}
    \caption{Accuracy and average training time per epoch for DAMP and DTKP-AM.}
    \label{fig:comparison_prov}
\end{figure}

\changed{We show the results of the provenance comparison experiments (RQ3) in Figure~\ref{fig:comparison_prov}. The graph on the top shows the accuracies achieved by each provenance over all tasks, while the bottom graph shows the average training time per epoch required for each provenance over all tasks.}

\section{Comparison with Tensor-based Neurosymbolic Frameworks}
\label{app:literature}

\changed{Systems like LYRICS~\citep{lyrics}, Logic Tensor Networks (LTNs)~\citep{ltn}, and Tensorlog~\citep{tensorlog} all have limited expressivity, which is one of the obstacles \tool{} aims to overcome. Specifically, they restrict the symbolic programs to first order logic and require users to specify low-level information such as how variables are grounded and what their domains are. They also restrict the symbols to be in the form of tensors and the user defined functions to consist of TensorFlow operations. These restrictions allow such systems to use TensorFlow to compile these programs into highly efficient computational graphs, but at the cost of expressivity. These frameworks also exclusively support simpler provenances and t-norms which are not sufficient for complex neurosymbolic programs.}

\changed{On the other hand, \tool{} allows the user to track tags for symbols which can be arbitrary Pythonic objects. \tool{} programs further allow the user to manipulate Distributions over such symbols using arbitrarily complex code which may not necessarily translate to a computational graph. As such, there is a fine balance between the probabilistic computations, that happen over a GPU, and the symbolic computations, that take place on a CPU, all while maintaining a mapping between the two. This fundamental design choice is also what allows \tool{} to be more expressive and flexible than existing systems. We also design \tool{} to be modular so that users can easily extend it to support new provenances and t-norms. As such, the t-norms used in LYRICS and LTN can be trivially added in a vectorized manner to \tool{}.}

\changed{For instance, assume the case of MNIST Sum-2, where `model` is the neural model. This is how it needs to be expressed in LTN:}
\begin{lstlisting}[style=PythonStyle, frame=none, escapechar=@@@@, basicstyle=\ttfamily\footnotesize]
### Predicates
Digit = ltn.Predicate.FromLogits(model, activation_function="softmax")
### Variables
d1 = ltn.Variable("digits1", range(10))
d2 = ltn.Variable("digits2", range(10))
### Operators
Not = ltn.Wrapper_Connective(ltn.fuzzy_ops.Not_Std())
And = ltn.Wrapper_Connective(ltn.fuzzy_ops.And_Prod())
Or = ltn.Wrapper_Connective(ltn.fuzzy_ops.Or_ProbSum())
Implies = ltn.Wrapper_Connective(ltn.fuzzy_ops.Implies_Reichenbach())
Forall = ltn.Wrapper_Quantifier(ltn.fuzzy_ops.Aggreg_pMeanError(),semantics="forall")
Exists = ltn.Wrapper_Quantifier(ltn.fuzzy_ops.Aggreg_pMean(),semantics="exists")


# mask
add = ltn.Function.Lambda(lambda inputs: inputs[0]+inputs[1])
equals = ltn.Predicate.Lambda(lambda inputs: inputs[0] == inputs[1])

### Axioms
@tf.function
def axioms(images_x, images_y, labels_z, p_schedule=tf.constant(2.)):
    images_x = ltn.Variable("x", images_x)
    images_y = ltn.Variable("y", images_y)
    labels_z = ltn.Variable("z", labels_z)
    axiom = Forall(
            ltn.diag(images_x,images_y,labels_z),
            Exists(
                (d1,d2),
                And(Digit([images_x,d1]),Digit([images_y,d2])),
                mask=equals([add([d1,d2]), labels_z]),
                p=p_schedule
            ),
            p=2
        )
    result_logits = axiom.tensor
    return result_logits
\end{lstlisting}
\changed{Note that the FOL semantics of the Real Logic language used in LTN requires the user to specify the tracking of the probabilities with the symbols denoted by the `digits*' variables.}

\changed{On the other hand, \tool{}'s design allows the user to write the same program in a more intuitive way:}
\begin{lstlisting}[style=PythonStyle, frame=none, escapechar=@, basicstyle=\ttfamily\footnotesize]
d1 = Distribution(model(img[0]), range(10))
d2 = Distribution(model(img[1]), range(10))

result_logits = GetProbs(Apply(d1, d2, lambda x, y: x + y))
\end{lstlisting}

\changed{\subsection{Optimizing Probabilistic Computations}}

\changed{Other works such as \citep{juice} and \citep{darwiche2020advance}, focus on solely on probabilistic computations rather than neurosymbolic frameworks. For instance, Juice~\citep{juice} is a Julia package for logic and probabilistic circuits, which is not designed to be integrated with deep learning frameworks. On the other hand, \cite{darwiche2020advance} focuses on variable elimination with applications to optimize tensor-based computation. It will be interesting to see how \tool{} can be integrated with such systems to further improve the scalability and efficiency of neurosymbolic learning, and will include a discussion on this in the revised manuscript. However, we still believe that \tool{}'s novelty lies in its design that allows for the seamless integration of general purpose neurosymbolic programs within deep learning frameworks, which is not addressed by the existing systems.}

\section{On Combinatorial Explosions}

\changed{The nature of the \textsc{Apply} and \textsc{ApplyIf} primitives imply the possibility of combinatorial ballooning of computations in cases where either the number of symbols is large or where there are several distributions over which the function is applied. This is indeed a fundamental challenge in neurosymbolic frameworks as a whole. \tool{} mitigates this by leveraging the Distribution class, which condenses symbols into a single collection stored in CPU RAM while maintaining tags as a GPU tensor ($b \times N \times T$, where $b$ is the batch size, $N$ is the number of symbols and $T$ is the shape of the tag). As shown in Figure~\ref{fig:mnist_sum_breakdown}, this approach reduces symbolic overhead by avoiding redundant evaluations for each batch sample, unlike frameworks like Scallop, where each sample in a batch is independently evaluated. While tag evaluations still involve all combinations across all samples in a batch, they are computed in a vectorized manner on the GPU.}

\changed{To see the effect of such computations even on larger experiments, we consider MNIST ProductN, where we multiply digits classified by the MNIST CNN as opposed to adding them in SumN. We show the per epoch training times in Table~\ref{tab:multn} for batch sizes of 64 and 256. In both cases, the \tool{} program is able to achieve high accuracies even for N=20 while running in reasonable amounts of time. The scaling gets even better for larger batch sizes (e.g. 256) since it increases the number of parallelized operations executed at any given time.}

\begin{table}[]
    \centering
    \caption{\changed{MNIST ProductN Training Epoch Times in Seconds.}}
    \begin{tabular}{ccccc}
    \toprule
    N & \multicolumn{2}{c}{B = 64} & \multicolumn{2}{c}{B = 256} \\
     & Time per Epoch (s) & Accuracy & Time per Epoch (s) & Accuracy \\
    \midrule
    4  & 11.42 & 0.96 & 8.92 & 0.97 \\
    8  & 12.55 & 0.95 & 9.15 & 0.95 \\
    16 & 27.45 & 0.94 & 15.71 & 0.89 \\
    20 & 36.59 & 0.92 & 18.73 & 0.85 \\
    \bottomrule
    \end{tabular}
    \label{tab:multn}
\end{table}

\section{The HWF Model}
\label{app:hwf}

\changed{We show the neurosymbolic model written in \tool{} for the HWF task along with the base neural model. In the HWF task, the neural model simply classifies each input image into 14 symbols: 10 digits and 4 operations.}

\begin{lstlisting}[style=PythonStyle, frame=none, escapechar=@@@@, basicstyle=\ttfamily\footnotesize]
class SymbolNet(nn.Module):
    def __init__(self):
      super(SymbolNet, self).__init__()
      self.conv1 = nn.Conv2d(1, 32, 3, stride = 1, padding = 1)
      self.conv2 = nn.Conv2d(32, 64, 3, stride = 1, padding = 1)
      self.fc1 = nn.Linear(30976, 128)
      self.fc1_bn = nn.BatchNorm1d(128)
      self.fc2 = nn.Linear(128, 14)
  
    def forward(self, x):
      x = self.conv1(x)
      x = F.relu(x)
      x = self.conv2(x)
      x = F.max_pool2d(x, 2)
      x = F.dropout(x, p=0.25, training=self.training)
      x = torch.flatten(x, 1)
      x = self.fc1(x)
      x = self.fc1_bn(x)
      x = F.relu(x)
      x = F.dropout(x, p=0.5, training=self.training)
      x = self.fc2(x)
      return F.softmax(x, dim=1)
\end{lstlisting}

\changed{This neural model is then used in the \tool{} program as follows:}
\begin{lstlisting}[style=PythonStyle, frame=none, escapechar=@, basicstyle=\ttfamily\footnotesize]
class HWFNet(nn.Module):
    def __init__(self):
      super(HWFNet, self).__init__()
  
      # Symbol embedding
      self.symbol_cnn = SymbolNet()
      self.operators = [("+", ), ("-", ), ("*", ), ("/", )]
      self.symbols = [ (str(i),) for i in range(10)] + self.operators

      self.db = torchql.Database()
  
    def forward(self, img_seq, img_seq_len):
      batch_size, formula_length, _, _, _ = img_seq.shape
      length = [l.item() for l in img_seq_len]

      inp = img_seq.flatten(start_dim=0, end_dim=1)
      symbol = self.symbol_cnn(inp).view(batch_size, -1, 14)
      
      def eval_formula(s):
        try:
          return eval("".join(s))
        except:
          return math.nan
        
      def concat_symbol(formula, symbol):
        if formula[-1] == "":
          return formula
        else:
          if not isinstance(symbol, tuple):
            symbol = (symbol,)
          formula += symbol
          if len(formula) %
            if formula[-2] in ["*", "/"]:
              eval_result = str(eval_formula(formula[-3:]))
              formula = formula[:-3] + (eval_result,)
          return formula
  
      def infer_expression(length, *symbols):
        res = symbols[0]
        for i in range(1, len(symbols)):
          res = Apply(res, symbols[i], concat_symbol)
        x = (Apply(res, eval_formula), )
        return x
  
      def reorg(symbols, lengths):
        distrs = []
        for i in range(symbol.shape[1]):
          if i < lengths:
            distrs.append(Distribution(symbols[i, :].view(-1, 14), self.symbols))
            if i %
              distrs[-1] = distrs[-1].filter(lambda s : s not in self.operators)
            else:
              distrs[-1] = distrs[-1].filter(lambda s : s in self.operators)
          else:
            distrs.append(Distribution(torch.ones(1, device=device), [("",), ]))
  
        res = (lengths, *distrs)
        return res
  
      q = torchql.Query("hwf", base="symbols").join("lengths") \
        .project(lambda symbols, lengths: reorg(symbols, lengths)) \
        .project(infer_expression, batch_size=batch_size)

      res = q(db, tensors={"symbols": symbol, "lengths": length}, disable=True).rows
      
      stacked = Distribution.stack(res)
      return GetProbs(stacked)
\end{lstlisting}

\changed{The \code{HWFNet} class is the neurosymbolic model. It takes in a sequence of images, \code{img\_seq}, and their lengths, \code{img\_seq\_len}. Note that within a single batch there may be image sequences of varying lengths. The neural model, \code{symbol\_cnn}, is used to classify each image in the sequence into one of the 14 symbols. Since we know that each number in the expression is a single digit, the \code{reorg} function is used to filter out relevant symbols based on their position in the sequence (operators in even places, digits in odd places). This function also pads sequences of smaller lengths with empty strings, written as Distributions with a single element and a probability of 1. Once reorganized, the \code{infer\_expression} function is used to infer the expression from the symbols. It does so by first concatenating Distributions using the \code{concat\_symbol} function, which also performs partial evaluations whenever possible. Once all the symbols are concatenated, the expression is evaluated using the \code{eval\_formula} function. The final expression is then returned as a Distribution. As a sidenote, while optional, we use the TorchQL~\citep{torchql} library to help write certain parts of the program. This shows the ease with which Distributions can be used with existing machine learning frameworks.}

\changed{For such a complex \tool{} program, using a simple provenance like DAMP proves insufficient for longer sequences since the tags of all possible combinations of symbols are collated into a single number. On the other hand, DTKP-AM is able to track the top-k proofs for each symbol, pruning out the less probable proofs. Furthermore, since each proof is a collection of \textit{input} symbols leading to a specific output, once the loss is calculated, gradients can be backpropogated directly to the input symbols that had the most influence on the output. On the other hand, the gradients may be distributed across all symbols in DAMP as it backpropogates through each intermediate computation regardless of their role in the computation of the output, resulting in slower convergence.}

\subsection{Writing HWF using LTNs}
\label{app:ltn-no-hwf}

The crux of the HWF program written in \tool{} relies on string concatenations and using Python's \code{eval} function.
In order to write HWF using LTNs, we need a different approach, since LTNs require all symbols to be grounded as tensors and all functions to be operations over those tensors.
While one can ground all the possible strings (representing $0, 1, \ldots, 9$ and operators $+, -, \times, /$) as real-value tensor encodings, one cannot execute functions like \code{eval} over those encodings  since it cannot be compiled onto the tensorflow computation graph.

This leaves us with the option of trying to write the HWF program as a probabilistic parser, as seen in Scallop. However, such a parser needs to be able to recursively parse arbitrarily long expressions, since HWF expressions are not of a fixed length even within the same task. This means it needs the ability to evaluate subtrees of the expression AST and produce intermediate results as new constants that can be populated within relations and operated over.

Based on our understanding of LTNs, the LTN paper does not provide any information on the dynamic creation of new constants or adding new domain elements at inference time, and instead focuses on examples and experiments where the domains are static and grounded prior to the evaluation of the logic formulae. We also spent considerable time perusing the official LTN repository, but could not find examples that could guide us towards writing an implementation of HWF.

We try to simulate the creation of a constant at inference time. In the following code, we attempt to write the program for MNIST Sum-3 by adding the first two digits in one LTN function, and then adding the result to the third digit, as follows (where the MNIST model is represented as \code{model}:

\begin{lstlisting}[style=PythonStyle, frame=none, escapechar=@@@@, basicstyle=\ttfamily\footnotesize]
### Predicates
Digit = ltn.Predicate.FromLogits(model, activation_function="softmax")
### Variables
d1 = ltn.Variable("digits1", range(10))
d2 = ltn.Variable("digits2", range(10))
d3 = ltn.Variable("digits3", range(10))
### Operators
Not = ltn.Wrapper_Connective(ltn.fuzzy_ops.Not_Std())
And = ltn.Wrapper_Connective(ltn.fuzzy_ops.And_Prod())
Or = ltn.Wrapper_Connective(ltn.fuzzy_ops.Or_ProbSum())
Implies = ltn.Wrapper_Connective(ltn.fuzzy_ops.Implies_Reichenbach())
Forall = ltn.Wrapper_Quantifier(ltn.fuzzy_ops.Aggreg_pMeanError(),semantics="forall")
Exists = ltn.Wrapper_Quantifier(ltn.fuzzy_ops.Aggreg_pMean(),semantics="exists")

# mask
add = ltn.Function.Lambda(lambda inputs: inputs[0]+inputs[1])
equals = ltn.Predicate.Lambda(lambda inputs: inputs[0] == inputs[1])

### Axioms
@tf.function
def axioms(images_x, images_y, images_w, labels_z, p_schedule=tf.constant(2.)):
    images_x = ltn.Variable("x", images_x)
    images_y = ltn.Variable("y", images_y)
    images_w = ltn.Variable("w", images_w)
    labels_z = ltn.Variable("z", labels_z)
    diagonal = ltn.diag(images_x,images_y, images_w, labels_z)
    formula = And(
        Digit([images_x, d1]),
            And(
                Digit([images_y, d2]),
                Digit([images_w, d3])
            )
    )
    exists = Exists(
                (d1,d2, d3),
                formula,
                mask=equals([add([add([d1, d2]), d3]), labels_z]),
                p=p_schedule
    )
    axiom = Forall(
            diagonal,
            exists,
            p=2
    )
    sat = axiom.tensor
    return sat

\end{lstlisting}

Here, rather than adding all three digits \code{d1}, \code{d2}, and \code{d3}, we write the following mask:
\code{equals([add([add([d1, d2]), d3])}. However, this line results in a typecheck error, since functions can only take terms, but \code{add([d1, d2])} returns a formula.

Given the lack of resources on such functionalities in LTN both in the paper, tutorials, and the examples in their repository, we were not able to write an implementation of HWF to evaluate \tool{} against.

\end{document}